\documentclass[Preprint,12pt,authoryear]{elsarticle}

\usepackage{subfigure}
\usepackage{setspace}
\usepackage{geometry}
\usepackage{framed,multirow}
\usepackage{latexsym}
\usepackage{bm}
\usepackage{amsmath}
\usepackage{booktabs}
\usepackage{xcolor}
\usepackage{color}
\usepackage{array}
\usepackage{algorithmic}
\usepackage{dblfloatfix}
\usepackage{verbatim}
\usepackage{lineno,hyperref}
\modulolinenumbers[1]

\journal{ISPRS Journal of Photogrammetry and Remote Sensing}

\usepackage{numcompress}

\begin{document}

\begin{frontmatter}
\title{Vehicle Detection of Multi-source Remote Sensing Data Using Active Fine-tuning Network}


\author{
 Xin Wu\textsuperscript{a}, Wei Li\textsuperscript{a}*, Danfeng Hong\textsuperscript{b}, Jiaojiao Tian\textsuperscript{b}, Ran Tao\textsuperscript{a}, Qian Du\textsuperscript{c}}

\address{
	\textsuperscript{a}School of Information and Electronics, Beijing Institute of Technology, 100081 Beijing, China, and Beijing Key Laboratory of Fractional Signals and Systems, 100081 Beijing, China (*corresponding author, e-mail: liwei089@ieee.org)\\
	\textsuperscript{b}Remote Sensing Technology Institute (IMF), German Aerospace Center (DLR), 82234 Wessling, Germany\\
	\textsuperscript{c}Department of Electrical and Computer Engineering, Mississippi State University, Mississippi State, 39762 MS, USA\\
}
\begin{abstract}
\textcolor{blue}{\textit{This is a pre-print of a paper accepted by ISPRS Journal Photogrammetry and Remote Sensing.}} \textcolor{red}{\textit{Please note that compared to the published version, we corrected several F1-Scores in Tables (marked in red) due to miscalculation.}}

Vehicle detection in remote sensing images has attracted increasing interest in recent years. However, its detection ability is limited due to lack of well-annotated samples, especially in densely crowded scenes. Furthermore, since a list of remotely sensed data sources is available, efficient exploitation of useful information from multi-source data for better vehicle detection is challenging. To solve the above issues, a multi-source active fine-tuning vehicle detection (Ms-AFt) framework is proposed, which integrates transfer learning, segmentation, and active classification into a unified framework for auto-labeling and detection. The proposed Ms-AFt employs a fine-tuning network to firstly generate a vehicle training set from an unlabeled dataset. To cope with the diversity of vehicle categories, a multi-source based segmentation branch is then designed to construct additional candidate object sets. The separation of high quality vehicles is realized by a designed attentive classifications network. Finally, all three branches are combined to achieve vehicle detection. Extensive experimental results conducted on two open ISPRS benchmark datasets, namely the Vaihingen village and Potsdam city datasets, demonstrate the superiority and effectiveness of the proposed Ms-AFt for vehicle detection. In addition, the generalization ability of Ms-AFt in dense remote sensing scenes is further verified on stereo aerial imagery of a large camping site.
\end{abstract}

\begin{keyword}
Multi-source, vehicle detection, optical remote sensing imagery, fine-tuning, segmentation, active classification network
\end{keyword}

\end{frontmatter}

\section{Introduction} \label{sec:In}
Recently, vehicle detection in remote sensing has been garnering growing attention and has been successfully applied in many applications, such as traffic management \citep{Arivalagan2015Vehicle,Karim2014Prototype}, and urban planning \citep{Wen2019,weng2018urban}. Visible (VIS) remote sensing imagery uses electromagnetic wave imaging at a wavelength of $400 \sim 760 $ nm, which visually reflects the true color and texture of vehicle objects. Inspired by the recent advancement of deep learning and availability of multi-source remote sensing, such as RGB, hyperspectral \citep{hong2018augmented}, multispectral \citep{weng2014generating}, and synthetic aperture radar (SAR) \citep{kang2020learning}, many state-of-the-art vehicle detection algorithms in remote sensing \citep{cheng2018learning,Yang2018Vehicle,Wu2018,Liu2015Fast, Nicolas2017Segment, Palubinskas2009Detection,6723710,cheng2020cross,wu2019fourier} have been developed. 

\begin{figure*}[!t]
    \begin{center}
        \includegraphics[width=\columnwidth]{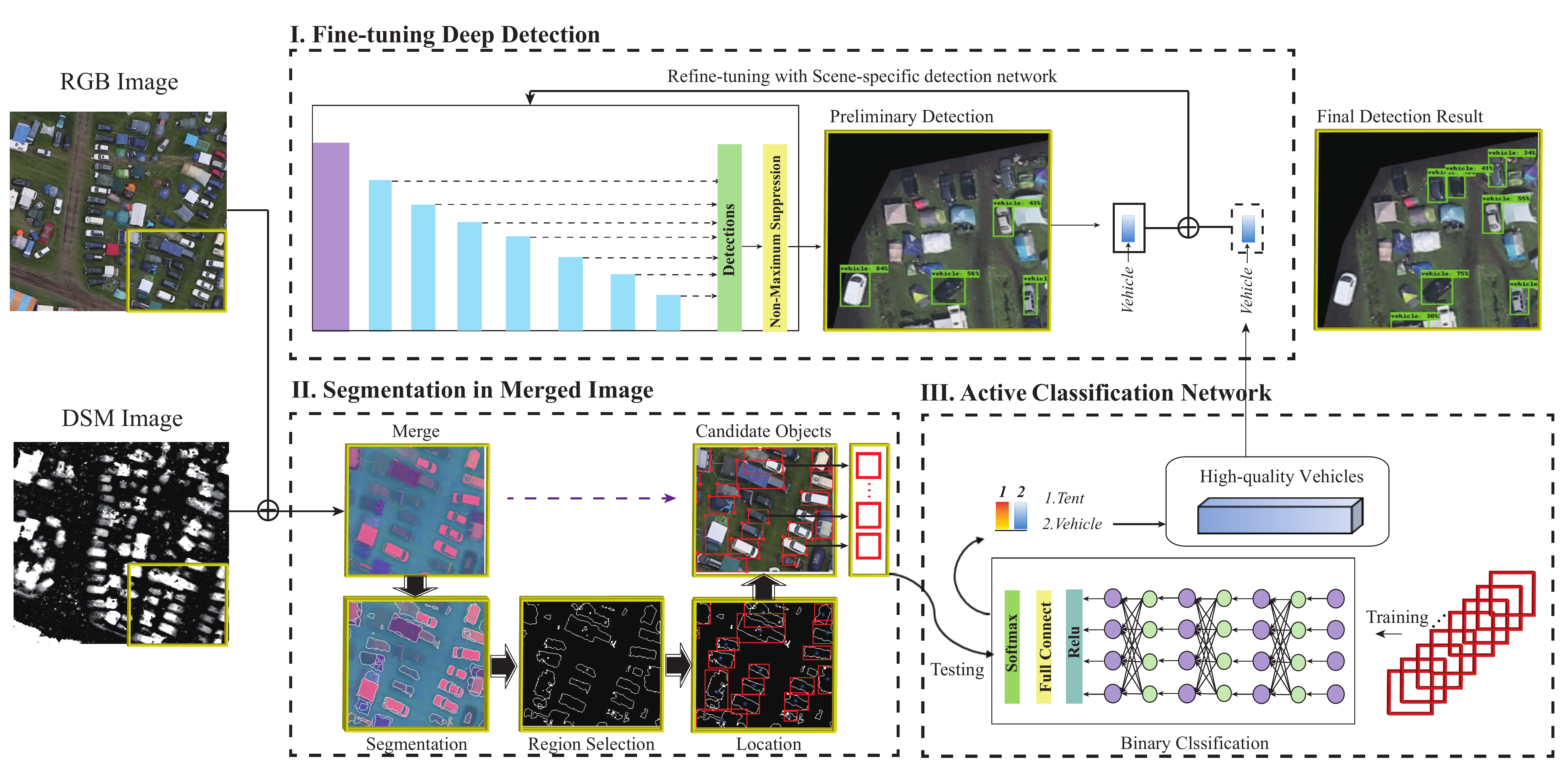}
        \caption{The overall flowchart of the proposed Ms-AFt, including automatic labeling and vehicle detection in complex remote sensing scene. More specifically, we first use a pre-trained detection network trained on other data sources (e.g., ImageNet), and the preliminary detection results of our unlabeled datasets can then be obtained by directly using the network. Next, an unsupervised object-based segmentation is performed using fused RGB and DSM images and potential objects can be screened out by the means of height information. Subsequently, these selected objects can be further classified into vehicles or non-vehicles using the branch III, i.e., active classification network. Finally, those high-quality vehicles with their location information are re-fed into the deep detection network to finely tune the network parameters.}
        \label{fig:outline}
    \end{center}
\end{figure*}

The success of using VIS images depends on the strong support of large-scale and accurate annotated datasets, e.g., NWPU VHR-10 dataset \citep{Cheng2014,Cheng2016,cheng2016survey} \footnote{\url{http://www.escience.cn/people/gongcheng/NWPU-VHR-10.html.}}, DOTA dataset \citep{xia2018dota} \footnote{\url{https://captain-whu.github.io/DOTA/dataset.html.}}, and DIOR dataset \citep{li2020object} \footnote{\url{http://www.escience.cn/people/gongcheng/DIOR.html.}}.
However, vehicles in VIS imagery are prone to have more complex deformation (see Fig. \ref{fig:outline}), due to its ``bird'' perspective. This inevitably increases the cost of manual annotation, thereby yielding a bottleneck. In order to solve this problem, a series of transfer learning \citep{Chen}, semi-supervised \citep{8519132}, weakly-supervised \citep{6991537} and even unsupervised methods
were successively developed to reduce the cost of annotation.  A novel weakly-supervised framework was proposed based on Bayesian principles for vehicle detection from VIS images \citep{6991537}. In \citep{8519132,Lin2016MARTA}, the generative adversarial network (GAN) has shown its superiority in data augmentation. GAN was developed to extract data distribution from unlabeled images in order to realize semi-supervised airplane detection \citep{8519132}. Furthermore, multiple-layer feature-matching generative adversarial networks (MARTA GANs) were designed to learn a representation using only unlabeled data to realize object classification \citep{Lin2016MARTA}. Subsequently, with a growing attention in limited or without labels, considerable work for object detection in remote sensing imagery has been reported in the literature \citep{inp,Cao2016Weakly}. Despite this, the detection ability using only a single data source still remains limited, due to the lack of feature diversity. For this reason, the joint use of multi-source data might be a good solution \citep{hong2019cospace}.

With the development of airborne sensors \citep{8648481}, the quantity and quality of remote sensing images have been greatly improved, which can make up, to some extent, for the limited performance of using a single data source. Some researchers have developed some work using multi-source data to achieve more accurate vehicle detection \citep{Schilling2018Detection}, but there is still room for improvement. Recently, digital surface models (DSM) have shown great potential in object detection. They can collect elevation information of objects and truly represents the ground ups and downs, yielding an effective separation of objects and background. There are two primary ways to generate DSM, namely LiDAR and photogrammetry. Among them, DSM obtained by LiDAR \citep{huang2019multi} has uniformity and high precision, but objects at the same location, especially for moving objects, rarely have LiDAR and aerial images at the same time. In addition, LiDAR equipment is expensive, so it is difficult to promote it in some practical applications. Thankfully, DSM obtained by photogrammetry \citep{Chai2016A}, that is reconstructing DSM directly on multi-view images or video data acquired by manned aircraft or UAV, can greatly reduce the cost and has been successfully applied in change detection, building extraction, auxiliary vehicle testing and other fields in recent years. A classification framework by combining morphological profiles, a spatial transformer network (STN) and a convolutional neural network (CNN) \citep{rasti2020feature} has been proposed for LiDAR-DSM data classification \citep{8480863}.

Inspired by the aforementioned investigation, we expect to further improve the detection performance of VIS images by taking the DSM as an auxiliary data source. Therefore, we do not extensively rely on the quality of DSM, as long as the used DSM can provide the performance improvement on the basis of that using VIS images. In other words, even though the DSM suffers from the effects that lead to the image degradation, e.g., slight move or missing of certain objects, tiny shift between VIS and DSM in the process of imaging or alignment, the quality of DSM is acceptable in our task. We have to admit, however, that the DSM with high quality can improve the detection results more effectively but also brings greater expense. As a trade-off, some existing datasets including DSM and VIS images, to a great extent, can meet our requirements.

In summary, VIS remote sensing imagery is sensitive to lighting, occlusion, and weather changes, so its ability to extract distinguishable features is still limited. DSM can work regardless of natural illumination. In order to effectively overcome the lack of distinguishability of a single sensor, data fusion is employed to make full use of the complementary information between multi-source images \citep{hong2019learnable}. Existing multi-source data-based methods \citep{Marmanis2016,Nie2017,Hen2018} use pre-fusion or post-fusion multiple sensor data under a supervised framework to improve the network's ability in feature learning rather than assisting in object detection. The main purpose of this paper is to realize vehicles auto-labeling to improve detection performance. Generally, it is very expensive and time-consuming to construct a large number of training samples for object detection. Therefore, it is meaningful but challenging to detect vehicles with an unlabeled training set.

To overcome these difficulties, a multi-source vehicle detection framework using an active fine-tuning network, Ms-AFt for short, is proposed. The proposed Ms-AFt network inherits the advantages of multi-source images, yielding more accurate labels of vehicle objects and higher detection performance. More specifically, the main contributions of this paper can be summarized as follow: 
\begin{itemize}
    \item To overcome the disadvantages of a single data source in vehicle detection, a strategy of using multi-source data is developed. Specially, DSM data without any label information are used to assist in locating ground objects in the segmentation branch to construct a vehicle candidate set. 
    \item Around multi-source data, the proposed Ms-AFt subtly integrates transfer learning, objects segmentation and active classification into a unified framework, ensuring vehicle auto-labeling with complementary information.
    \item In the proposed Ms-AFt, an active classification network is designed, which can effectively screen high quality vehicles to increase the diversity and number of training sets with progressive improvement in vehicle detection.
\end{itemize}

\begin{figure*}[!t]
    \begin{center}
        \includegraphics[width=\columnwidth]{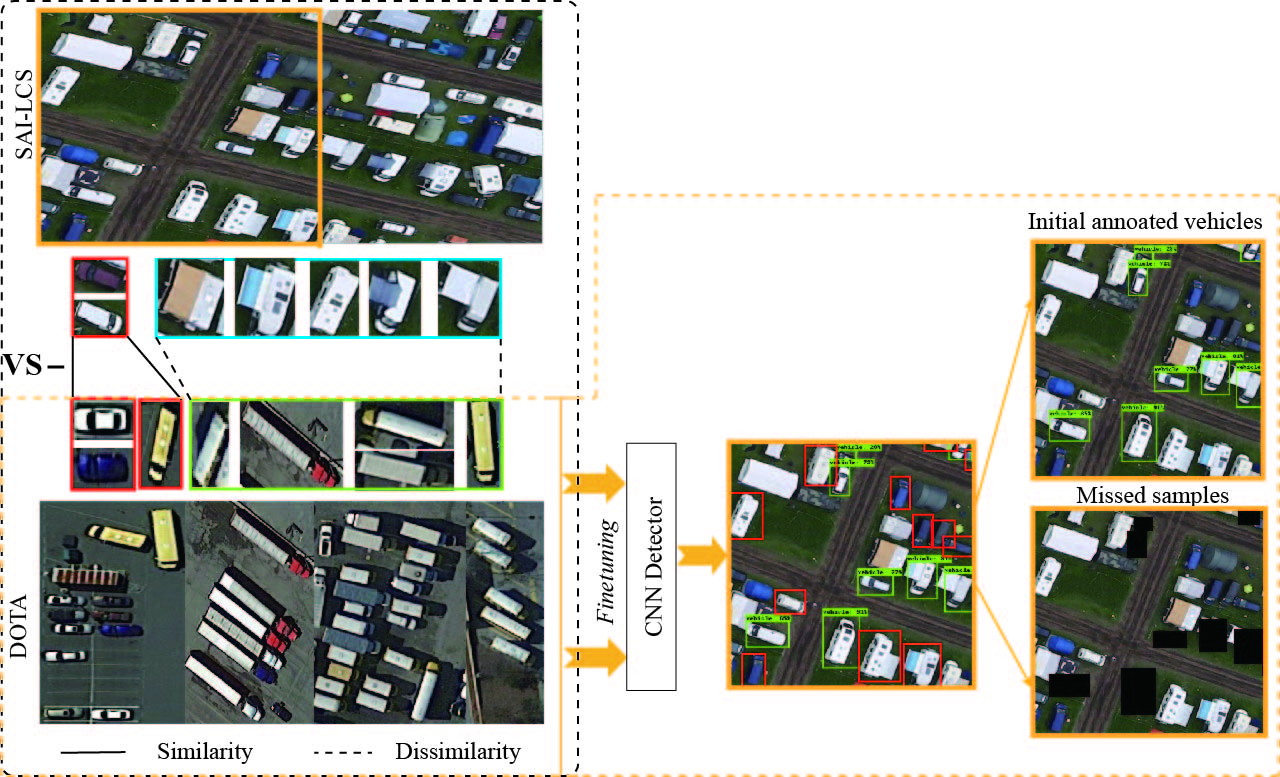}
        \caption{Illustration of the fine-tuning branch. Vehicle samples are shown within the dashed lines of a DOTA and SAI-LCS.}
        \label{fig:fine-tuning}
    \end{center}
\end{figure*}

The remainder of this paper is organized as follows. Section \ref{sec:Me} describes the three branches of the proposed unlabeled vehicle detection framework, including transfer learning, object segmentation and active classification. Section \ref{sec:Exp} validates the proposed framework and reports detection results, as compared with six competitive networks. Section \ref{sec:Con} draws the conclusions and briefly discusses future work. 

\section{Proposed Vehicle Detection Framework }\label{sec:Me}

As illustrated in Fig. \ref{fig:outline}, the proposed Ms-AFt network consists of three branches. They are the fine-tuning branch, segmentation branch, and active classification branch. In the first branch, VIS images are fed into a pre-trained network to generate the first part of the vehicles training sets. To enrich the diversity of vehicle categories, the multi-source based segmentation branch is designed to generate additional candidate object sets. Then, an active classification network with the classic ResNet18 architecture is used to classify different objects, e.g., buildings, tents, vehicles, and further screen out the high-quality vehicles from candidate object sets. These selected candidates can be augmented to the vehicle training set and used to finely tune the designed detection network for the final vehicle detection.

\subsection{Fine-tuning Branch}\label{sec:A}

As the data used in practice are usually unlabeled, we attempt to answer the following question: \textit{How can we  implement vehicle detection similar to world-class images?} In transfer learning, a reference dataset and a target dataset are sufficiently similar, and the reference dataset is often much larger than the target dataset. Therefore, a pre-training network is introduced and fine-tuned - only the Dataset for Object deTection in Aerial (DOTA) images with the annotation information containing at least one vehicle object while keeping the other configurations of the network the same, yielding a detection result for the unlabeled dataset. DOTA is an aerial image dataset with 15 categories of objects collected by Wuhan University, Huazhong University of Science, and Technology and the German Aerospace Center. This dataset contains 2806 aerial images from different sensors and platforms and 188,282 fully annotated objects. Each image is of size about $4000\times 4000$ pixels. So far, it is the largest open-sourced labeling dataset in the field of remote sensing for object detection. 

In terms of limited datasets, computing resources, and time, using a pre-trained model is the best choice. Existing work \citep{Lin2016MARTA} has confirmed experimentally that a model trained from scratch cannot surpass the fine-tuned model based on a pre-trained network. Some representative pre-trained networks, which are listed in Section \ref{sec:Exp_Dis} and used in this paper are mainly trained by a large visual dataset (e.g., Imagenet, COCO). For the network fine-tuning, we first separate the vehicles from the DOTA dataset and build a small DOTA vehicle dataset, and then the above pre-trained networks are fine-tuned. Except that the output class number and the learning rate are adjusted to 2 and 0.001, respectively, other network parameters remain unchanged during network fine-tuning. For simplicity, all the layers, rather than freezing several layers or redesigning the network, are retrained. In addition, the final training set of the detected dataset includes all DOTA vehicle samples in order to prevent overfitting.

However, detection performance is closely related to the similarity between different datasets, which limits the generalization performance. Fig. \ref{fig:fine-tuning} takes the Stereo Aerial Imagery of a Large Camping Site (SAI-LCS) dataset constructed by the German Aerospace Center as an example to show this dataset and DOTA containing vehicle samples. As we can see, the SAI-LCS has a more complex background and a wider variety of vehicles than DOTA. The background of the images in SAI-LCS is a grass lawn, which greatly reduces the difference between objects and background. In addition, there are many types of vehicles in the SAI-LCS, including recreational vehicles (RVs), suspected RVs, etc, which are not in DOTA, and the objects are densely arranged and even have adhesions. The large-scale missed alarm in the detection results as shown in Fig. \ref{fig:fine-tuning} also confirms the above statement, so it is not sufficient for effective vehicle detection in the SAI-LCS by simply using DOTA vehicle images for fine-tuning. 

\subsection{Object Segmentation Branch}\label{sec:B}

In view of the limitations of network fine-tuning, a feasible solution to improve vehicle detection performance for unlabeled datasets is to automatically increase the diversity of vehicles while expanding the number of training sets. In other words, we firstly separate objects from ground in the image, construct non-similar vehicle candidate sets, and then use the active vehicle classification network to select high quality vehicles. Among them, object separation is discussed in this section, and the active classification will be discussed in Section \ref{sec:C}. 

The segmentation-based method is one of the most effective methods to separate objects from the ground. The classic pixel-based segmentation method \citep{ZANOTTA2018162} is suitable for VIS images, which are disturbed by illumination, occlusion, shadow, background clutter, and other factors. In addition, excessive reliance on spectral information can also result in a large number of fractured areas, which inevitably reduces the quality of segmentation, especially for long-range aerial images. Here, DSM is considered to compensate for the disadvantages of VIS. It covers the terrain and other surface information except for the ground, expressing the fluctuation of the ground most realistically. 

In order to make full use of multi-source image information to improve the accuracy of ground object positioning, a scalar weighted fusion strategy, especially for both sensors with high quality, is employed. For the segmentation method, object rigid structure and adopt superpixel segmentation \citep{hong2020invariant} are utilized to achieve separation of dense ground objects. Superpixel can remove redundant information and fit edges and its processing speed is more than ten times faster, so it is more suitable for positioning densely arranged objects. Fig. \ref{fig:outline} (see part II) shows the flowchart of object segmentation. The main steps are summarized as follows.

\textit{1) Superpixel Segmentation and Clustering}: simple linear iterative clustering (SLIC) \citep{achanta2012slic} is chosen for image superpixel segmentation. The superpixel has good compactness and boundary fit with fast computation speed, which well maintains the object contour \citep{hong2020invariant}. SLIC iterates around distance measurement. Note that the number of superpixels in SLIC is a to-be-set hyperparameter. In our case, the parameter is empirically and experimentally determined to be around 2000. If $D$ between the two cluster centers is less than a certain threshold, superpixels and corresponding adjacency matrices of each cluster can be returned. As for the spatial aggregation, density-based spatial clustering of applications with noise (DBSCAN) \citep{ester1996density} is considered, which does not need to specify the number of clusters. Given the neighborhood parameters $\left( {\varepsilon, MinPts} \right)$, the clustering result is deterministic, and it can also solve the special situation of data distribution.

\textit{2) Region Selection and Localization of Ground Objects}: to effectively eliminate the impact of objects with higher elevation information in the image (e.g., trees, high-rise buildings, etc.) for the final vehicle separation, the DSM image is thresholded to keep it within a certain height interval, and each cluster value in this image is defined as its average height. In detail, given the to-be-selected candidates, denoted as $\mathbf{z}_{i}, \; i = 1,....,p$, from the step 1), we expect to screen out the regions ($R_{j}(\mathbf{z}_{i}),\; j = 1,...,q$, where $p\gg q$) with a proper size via a \textit{hard threshold} (th), e.g.,

\begin{equation}
\label{eq1}
\begin{aligned}
       R_{j}(\mathbf{z}_{i})=
       \begin{cases}
       1, & \rm{if} \; \rm{area}(\mathbf{z}_{i})< th, \\
       0, & \rm{otherwise},
       \end{cases}
\end{aligned}
\end{equation}
where the symbol $\rm{area}(\bullet)$ is defined as the area computation of the $\bullet$ candidate. Using Eq. (\ref{eq1}), these candidates can be effectively categorized into two groups by a given threshold, according to the order of the area size. Once the selected ground objects are determined, the morphological filtering with dilation and erosion is used to further smooth their edges.

\textit{3) Vehicle Candidate Set Generation}: the purpose of vehicle detection is classification and positioning, but the diversity of vehicle types in aerial image datasets inevitably increases the complexity of strict vehicle positioning. Here, the smallest circumscribed rectangle of the selected region is constructed, then a corner detector is employed to obtain the diagonal location information of objects in each selected region, and the horizontal bounding box location of the object is implemented.

\subsection{Actively Classification Branch} \label{sec:C}

\begin{figure*}[!t]
    \begin{center}
        \includegraphics[width=\columnwidth]{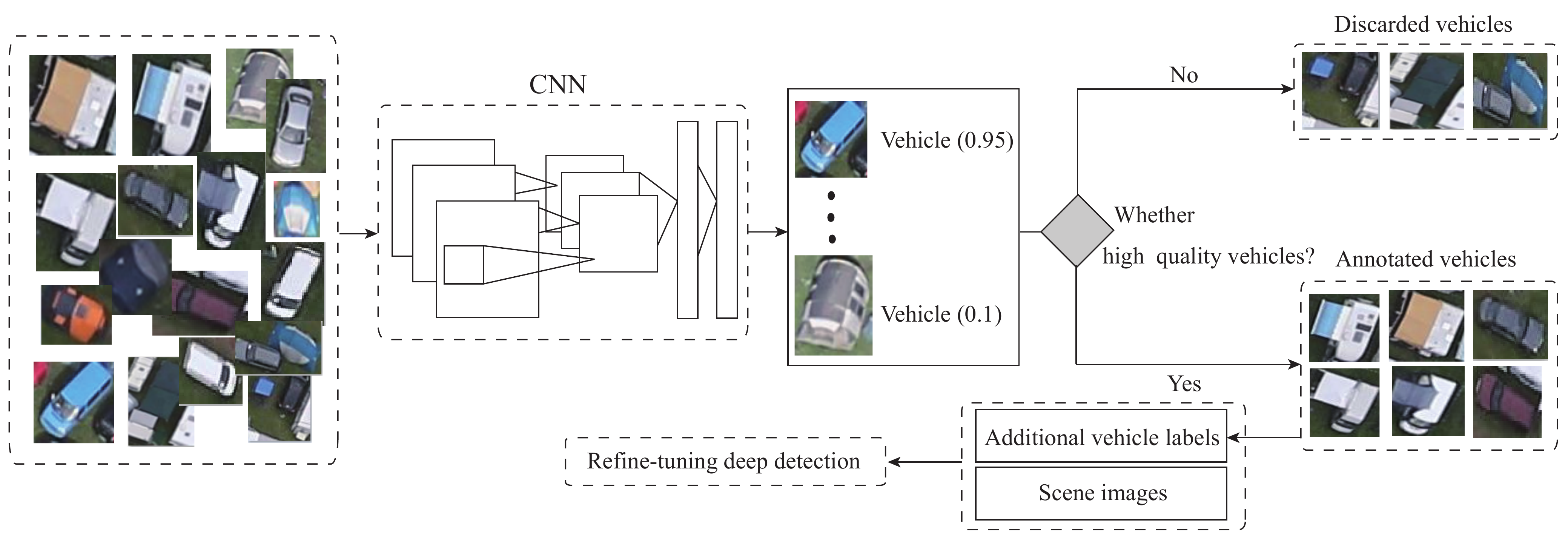}
        \caption{Illustration of the designed active classification branch. The pipeline includes CNN and model initialization, detection training set updating, picking high-confidence samples selection and labeling, where the arrows indicate the workflow. }
        \label{fig:active_cls}
    \end{center}
\end{figure*}

\begin{figure*}[!t]
    \begin{center}
        \includegraphics[width=\columnwidth]{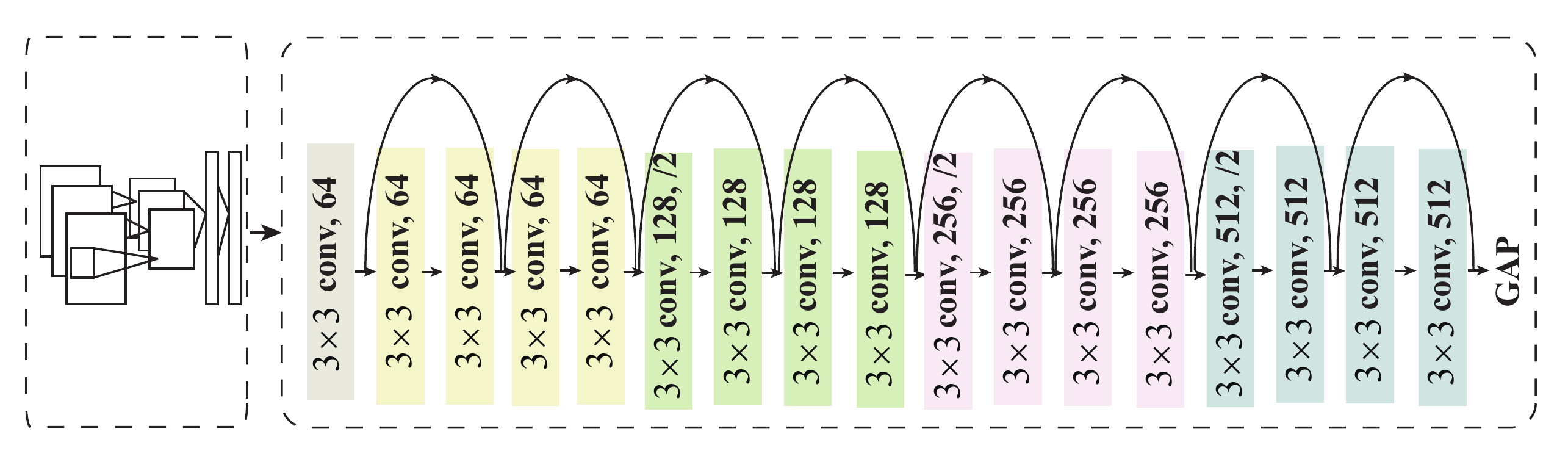}
        \caption{The detailed framework of CNN for active classification network. }
        \label{fig:active_net}
    \end{center}
\end{figure*}
\begin{table*}[!t]
    \footnotesize
    \centering
    \caption{Classification performance comparisons of four different classifiers. The best results are highlighted in bold.}
   \begin{tabular}{c c c c c}
        \toprule
        Dataset & Method & Precion & Recall & F1-Score\\ \hline \hline
        \multirow{4}{*}{ISPRS Vaihigen} & VGG-19 & 0.93 & 0.67 & 0.78  \\ 
        & GoogleNet &  0.95 &  0.76 &  0.84\\ 
        & Cascadenet18 & 0.95 & 0.74 & 0.83  \\
        & ResNet18 &  \bf 0.96 &  \bf 0.86 &  \bf 0.91 \\ \hline
        \multirow{4}{*}{ISPRS Potsdam} & VGG-19 & 0.95 & 0.78 & \textcolor{red}{0.86}  \\ 
        & GoogleNet &  0.97 &  0.85 &  0.91\\ 
        & Cascadenet18 & 0.96 & 0.82 & 0.88  \\
        & ResNet18 &  \bf 0.99 &  \bf 0.90 &  \bf \textcolor{red}{0.94} \\ \hline
        \multirow{4}{*}{DLR SAI-LCS} & VGG-19 & 0.92 & 0.52 & \textcolor{red}{0.66} \\
        & GoogleNet &  0.96 &  0.70 &  0.81\\ 
        & Cascadenet18 & 0.96 & 0.58 & 0.72  \\
        & ResNet18 &  \bf 0.97 &  \bf 0.76 &  \bf 0.86 \\
        \bottomrule
    \end{tabular}
    \label{tab:classification}
\end{table*}

As the vehicle categories are more than one, effective selection of high quality vehicles from the vehicle candidate set are discussed in Section \ref{sec:C}. An active classification network inspired by active learning is employed. Different from the classical active learning method with manual labeling by related experts, the active classification branch appoints the ResNet18 network as the active selection strategy to automatically learn the high quality vehicles from candidate objects. This process is called ``active classification''. Fig. \ref{fig:active_cls} shows the flowchart for this branch, and its detailed steps are as follow.

\textit{Step1: Classification Training Set Construction}

There are three parts of vehicle samples in the classification network, namely the vehicles in DOTA, the detected vehicles in target dataset by Section \ref{sec:A}, and the manually labeled vehicles in this section. In comparison, the third part of the classification training set plays a decisive role in the whole classification branch. It mainly focuses on the characteristics of vehicles, including various scales, directions, shapes, and so on, which can greatly improve the vehicle classification performance. In the experiments, it is necessary to know the average size of vehicles (objects) for better training the classification network.

\textit{Step2: High Quality Vehicles Selection}

Four pre-trained classification networks are compared, namely, VGG-19 \citep{Karen2014VGG}, GoogleNet \citep{Szegedy2016Inception}, Cascadenet18 \citep{Pang2018Cascade}, and ResNet18 \citep{He2015ResNet}. Table \ref{tab:classification} lists the performance of these four classification networks. For a fair comparison, the parameters of each network are optimized. Overall, the classification performance of ResNet18 is the best. The possible reason is that each layer of the ResNet with shortcut connection has a survival probability, and would be discarded according to actual needs, which makes the ResNet always yield the best performance, but the other three networks are not adjustable. In addition, ResNet can ease the speed of performance descent when the resolution of the input image is fine enough. Specifically, the recall of the ISPRS dataset is much higher than that of the DLR data set because the vehicle types of the ISPRS dataset are more general and the existing public vehicle samples are sufficient to cover.

By following the above analysis, ResNet18 is selected as the classification network due to the fact that the resolution of vehicles is relatively small. The detailed architecture of the classification network is shown in Fig. \ref{fig:active_net}. Among them, the GAP in the network is the global average pooling layer. Given a training set $S=\left\{ {\left( {\mathbf{x}_1, y_1} \right), \cdots ,\left( {\mathbf{x}_N, y_N} \right)} \right\}$ with $N$ samples and $K$ categories, where $\mathbf{x}_{i}$ and $\mathbf{y}_{i}$ denote the $i$-th sample and its corresponding label, respectively, we then define the following cross-entropy loss $L_{C}$ for a binary classification task.
\begin{equation}
\begin{aligned}
      L_{C}=\frac{-1}{N}\sum_{i=1}^{N}[y_{i}{\rm log}\mathbf{a_{Soft}}_{i}+(1-y_{i}){\rm log}(1-\mathbf{a_{Soft}}_{i})],
\end{aligned}
\end{equation}
where $\mathbf{a_{Soft}}$ denotes the feature outputs after going through the softmax layer.

\subsection{Analysis on Proposed Ms-AFt} 

The recent success of CNN-based architecture brings the power in vehicle detection, owing to sufficient well-annotated samples \citep{Yang2018Vehicle,8694857,8755462,Schilling2018Detection}. However, costly manual labeling makes it difficult to acquire a large number of labeled samples in practice, leading to the poor detection performance of the previous network-based methods, e.g., FCN \citep{Schilling2018Detection}. Therefore, it is a feasible solution to build an effective auto-labeling method to expand the number and categories of training samples. Also, the introduction of multi-source data, e.g., DSM with height information can roughly locate and label the ground objects inferred, can further improve the accuracy of object auto-labeling and finally detection performance.
\begin{figure*}[h!]
    \begin{center}
        \includegraphics[width=\columnwidth]{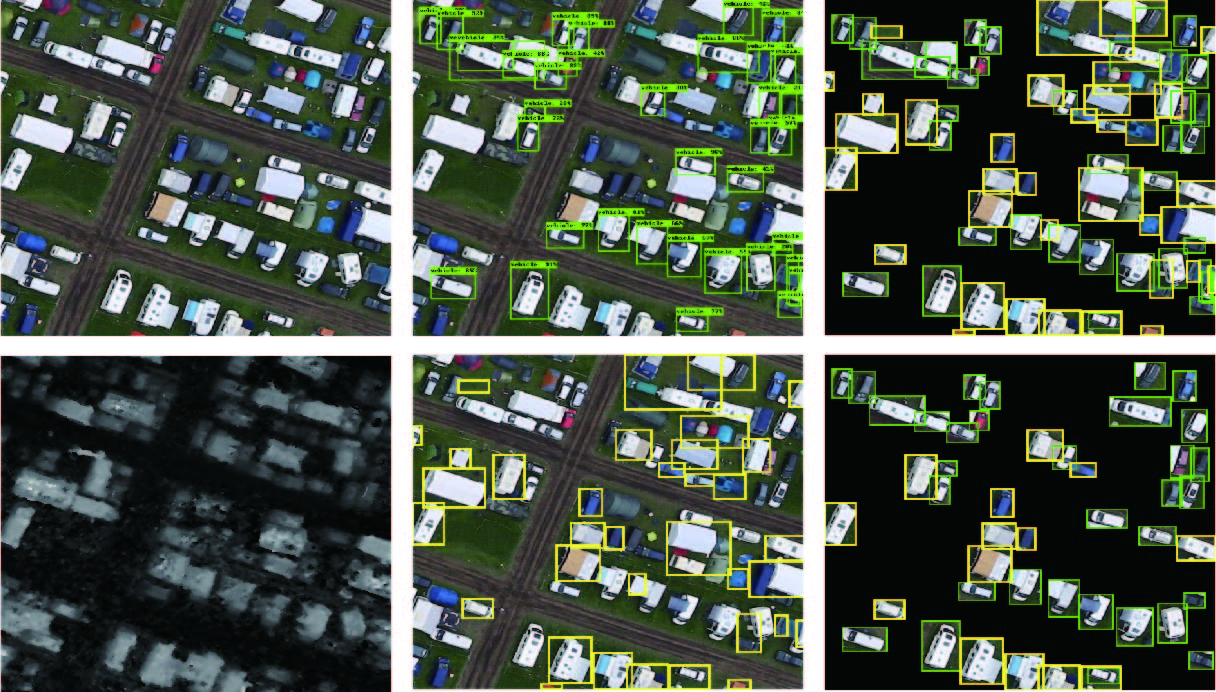}
        \caption{Visualization of our vehicle auto-labeling workflow for the SAI-LCS dataset. First column: VIS data (up) and DSM image (down). Second column: the auto-labeling results of fine-tuning branch (up) and segmentation branch (down). Third column: the auto-labeling objects mask (up) and high quality vehicles results (down).}
        \label{fig:visual}
    \end{center}
\end{figure*}

\begin{figure*}[h!]
    \begin{center}
        \includegraphics[width=\columnwidth]{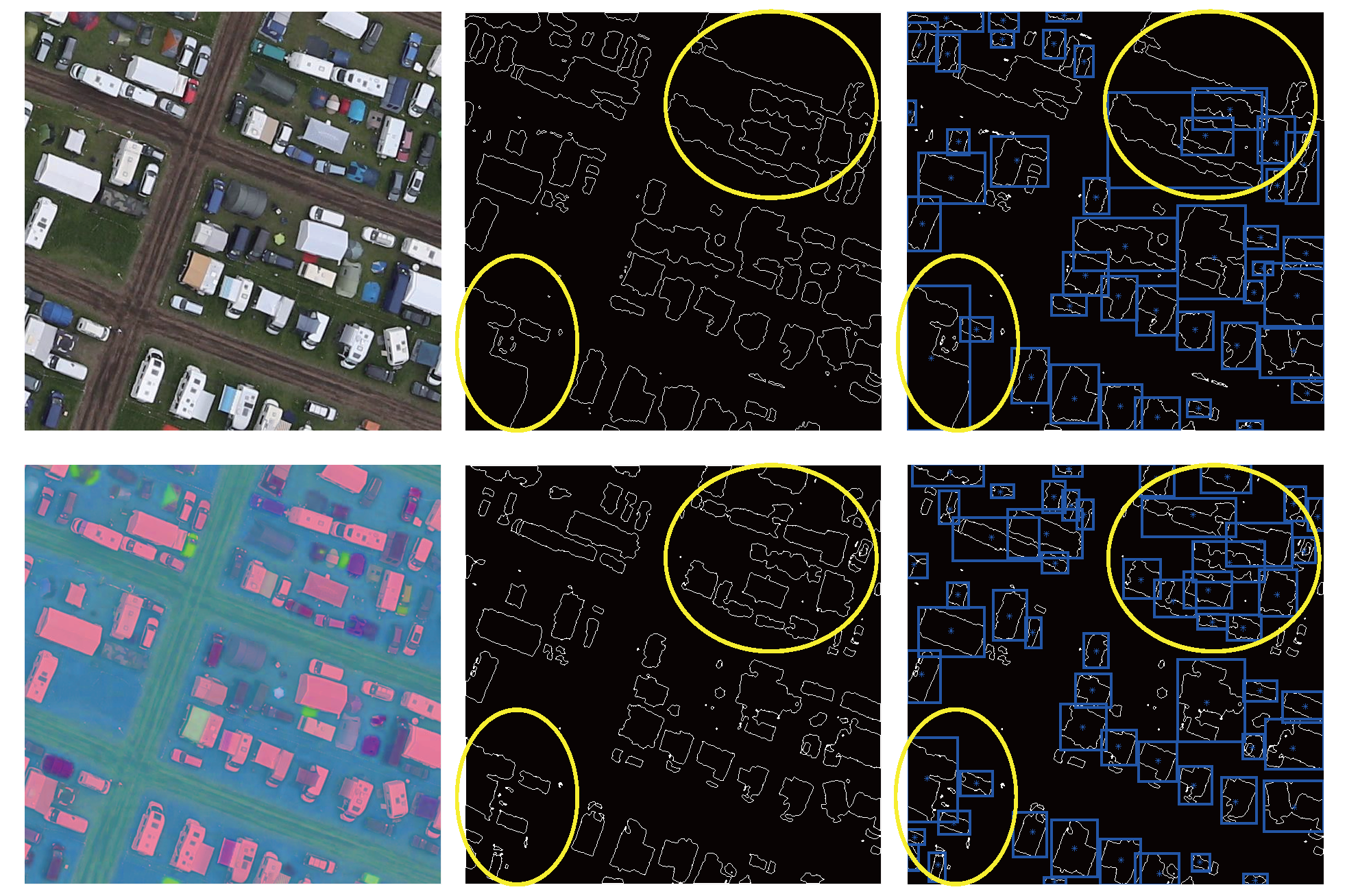}
        \caption{An example image for auto-labeling in the segmentation branch. The first row and second row shows the results of VIS and merged image, respectively. The yellow ellipse points to two distinct discrimination in the visualization. }
        \label{fig:seg}
    \end{center}
\end{figure*}

The proposed Ms-AFt framework mainly focuses on vehicle auto-labeling. Except for the detection class and learning rate, the classic network architecture used in this paper is not revisted or updated. Fig. \ref{fig:visual} illustrates a visual example of the auto-labeling process. The joint use of three branches in the proposed Ms-AFt network can effectively improve the detection performance of vehicles by means of multi-source data. In the first branch, VIS images are sent to a pre-trained network to generate the first part of the vehicle training sets for an unlabeled dataset. To cope with the diversity of vehicle categories, the multi-source based segmentation branch is designed to generate additional candidate object sets. Then, an active classification network with the classic ResNet18 network as an automatic selection strategy is constructed to confirm the valid category of unknown high quality vehicles from candidate object sets, and to augment the vehicle training sample set for final vehicle detection. In addition, the Ms-AFt framework is based on the constraint that: 1) the labeled reference vehicle dataset is complex enough; in other words, it can approximate the real world reflection; 2) the VIS and DSM are acquired at the same time. Specifically, the multi-source data used in the segmentation branch is mainly to improve the accuracy of ground objects positioning, rather than improving the discriminability of the network. The DSM does not participate in the comparison because it inevitably results in poor edge segmentation due to the impact of filling. Fig. \ref{fig:seg} illustrates an example of segmentation results. In the yellow ellipse area, it can be visually seen that the multi-source image helps to reduce adhesion between regions. In addition, the DSM of dark vehicles can effectively make up insufficient contrast between vehicles and ground in the VIS image, improving the accuracy of its positioning.

To reduce the complexity of ground object positioning, a generalized bounding boxes positioning method, named horizontal bounding boxes (HBB), is utilized. A common description of HBB is $\left({x_c,y_c,w,h}\right)$, where  $\left({x_c,y_c}\right)$ is the center location, $w, h$ are the width and height of the bounding box, respectively. In addition, regional connectivity is examined to ensure that only one object is included in each selected area.

\section{Experimental Results and Analysis} \label{sec:Exp}

\subsection{Data Description} \label{sec:Data}

The performance of the proposed active fine-tuning network are quantitatively and qualitatively evaluated on three representative multi-source datasets: two from the ISPRS 2-D semantic segmentation baseline dataset (Vailingen village dataset\footnote{\url{http://www2.isprs.org/commissions/comm3/wg4/2d-sem-label-vaihingen.html.}} and Potsdam city dataset\footnote{\url{http://www2.isprs.org/commissions/comm3/wg4/2d-sem-label-potsdam.html.}}) and the Stereo Aerial Imagery of a Large Camping
Site (SAI-LCS) dataset \citep{Tian2018}. It should be noted that this paper aims at improving the detection performance using multi-source remote sensing data, e.g., VIS and DSM images. Therefore, this requires a good alignment between multi-source data. The three used datasets are generated by the photogrammetric way. This means that the two products (VIS and DSM) are actually produced from a single data source, e.g., aerial imagery. Such method, to a great extent, can meet the alignment requirement. More details regarding these datasets are given as follows.

\subsubsection{\textit{ISPRS 2-D Semantic Segmentation Baseline Dataset}}

\textit{DSM Generation Process:} DSMs for ISPRS 2-D Semantic dataset is by photogrammetry. This dataset consists of very high resolution orthophoto (TOP) tiles and the corresponding DSM derived from dense image-matching techniques. Both areas cover urban scenes. Among them, the TOP image is generated using Trimble INPHO OrthoVista, and the DSM is produced by dense image matching using Trimble INPHO 5.3 software.

\textit{Experiment Data Introduction:} \textbf{1) Vaihingen Village Dataset:} Vaihingen dataset consists of 33 differently sized areas and its ground sampling distance is 9 cm. The TOP of this dataset is an 8-bit TIFF file with three bands (near infrared, red and green channels available). There are occlusions and shadows in the vehicle parked area for each image. These areas constitute an opaque wall, which greatly increases the difficulty of vehicle detection, especially for dark vehicles. In the following experiment, the vehicle training set contains 28 aligned VIS and DSM scene images with approximately 700 vehicles. The testing set has 5 VIS scene images with 148 manually labeled vehicle samples; \textbf{2) Potsdam City Dataset:} this dataset includes 38 differently sized areas and its ground sampling distance is 5cm. The TOP of this dataset is an 12-bit TIFF file with four bands (near infrared, red, green and blue channels available). Overall, there are few overlapped objects (or occlusion) in the parking area of the image scene. Compared with the other two data sets, vehicle detection from this image is relatively simple. In the experiment, about 6,000 vehicles contained in 33 aligned VIS and DSM scene images are used as the training set. The testing set has 5 VIS scene images with 1874 manually labeled vehicle samples.

Since the main focus of this paper is to realize vehicles auto-labeling for vehicle detection, the ground truth in this dataset is only used for testing.

\subsubsection{\textit{SAI-LCS Dataset}} 

\textit{DSM Generation Process:} DSMs for SAI-LCS dataset were collected by photogrammetry. Firstly, optical imagery was collected, acquired by the optical 4K camera system on the German Aerospace Center (Deutsches Zentrum für Luft- und Raumfahrt, DLR) research helicopter BO 105 (DLR (CC-BY 3.0)) at a height of 600m above the ground.  The ground sampling distance (GSD) of images was around 11 cm. With three cameras on board, the 4K camera system is able to capture the multi-view imagery with 90\% overlap along-track and 60\% overlap across the track. DSMs were reconstructed by a Structure from Motion (SfM) technique, and literature \citep{Tian2018} described these in detail.

\textit{Experiment Data Introduction:} This SAI-LCS dataset is a subset from the aerial imagery of a large camping site in northern Germany, which comprised an area of $1.0\rm{km}\times 1.5\rm{km}$. In the experiment, the vehicle training set contains 40 aligned VIS and DSM scene images with about 900 vehicles. The testing set has 60 VIS scene images with 2313 manually labeled vehicle samples.

\subsection{Experimental Setup}

In the experiment, the input image of detection is always resized to a fixed shape of $608\times608$ pixels. For avoiding object splitting, the step size is set to $304$ and the testing result only retains IOU (Intersection over Union) $\textgreater0.7$. In the active deep classification network, the resolution  of the image in the vehicle candidate set is rescaled and resampled to $60\times 60$ pixels. To prevent over-fitting classification,  all the images in the training set are rotated from $0^{\circ}$ to $270^{\circ}$ in steps of $90^{\circ}$. In addition, images are transformed to the HSV color space to improve the network's robustness to change of illumination.
\begin{figure*}[h!]
    \begin{center}
        \includegraphics[width=\columnwidth]{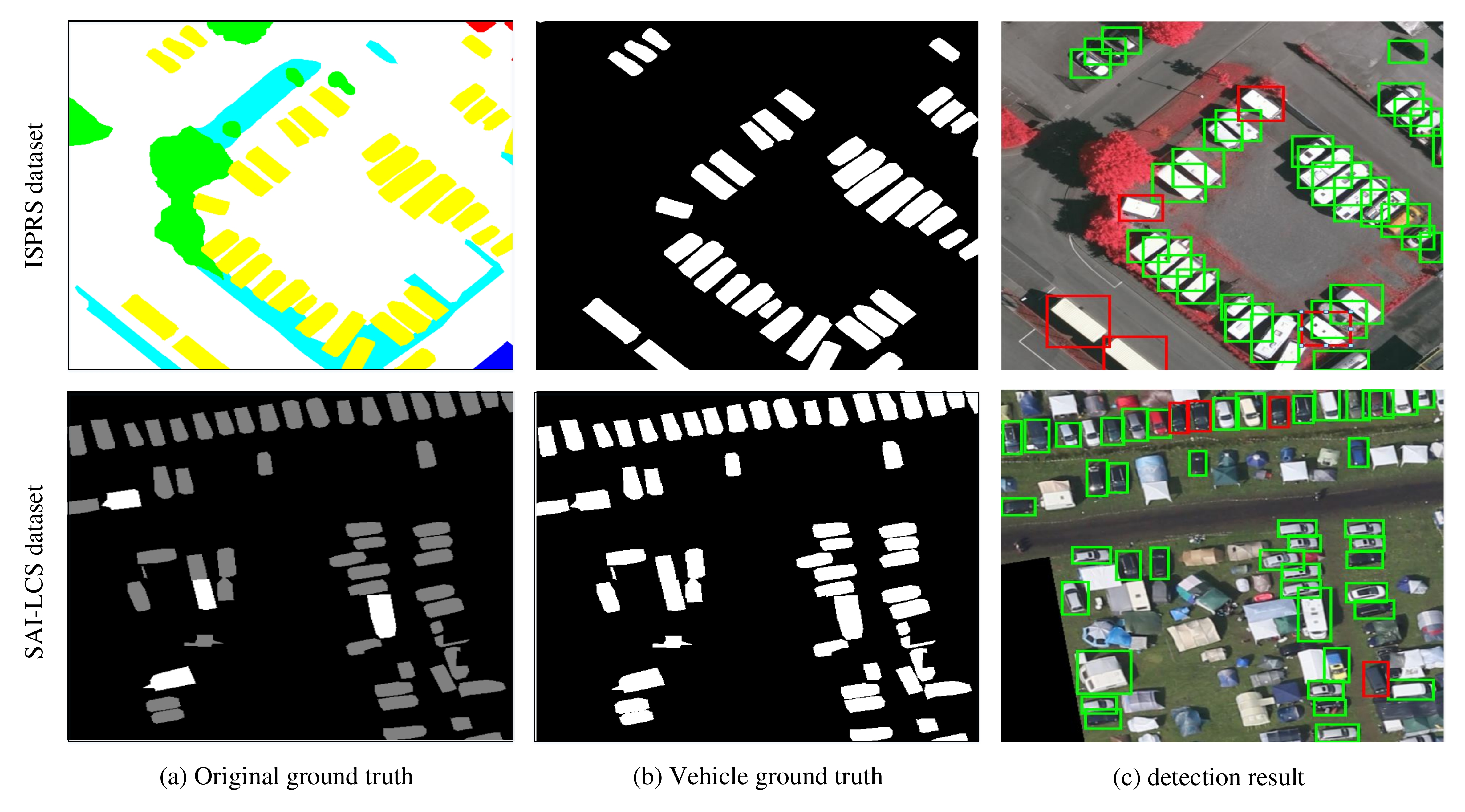}
        \caption{Qualitative evaluation of the experimental datasets. (a) is the given original ground truth of the scene including different categories. (b) is the corresponding vehicle ground truth selected from (a). (c) shows our vehicle detection results, where a green box denotes correct detection, a red box denotes the leak detection.}
        \label{fig:gt}
    \end{center}
\end{figure*}

Five commonly-used criteria, \textit{Precision-Recall Curve (PRC)}, \textit{Average Precision ($AP$)}, \textit{Precision (P)}, \textit{Recall (R)}, and \textit{F1-score} are adopted to quantify detection accuracy. Among them, $AP$ is a global indicator measured by the area under the PRC. The higher the AP value, the better the performance. The \textit{Precision (P)} is computed by $\frac{TP}{TP+FP}$, and the \textit{Recall (R)} rate is $\frac{TP}{TP+FN}$, TP, FP, and FN denote true positive, false positive, and false negative, respectively. Moreover, the \textit{F1-score} and \textit{AP} can be computed by using the following equations
\begin{equation}
\begin{aligned}
       F1= \frac{2*P*R}{(P+R)},
\end{aligned}
\end{equation}
and 
\begin{equation}
\begin{aligned}
       AP= \sum_{k=1}^{n}P(k)\Delta R(k),
\end{aligned}
\end{equation}
where $P(k)$ and $\Delta R(k)=R(k)-R(k-1)$ denote the precision and recall difference, respectively, when using the $k$-th threshold.

All the deep learning experiments are implemented with the Tensorflow framework and carried out by a PC with an Intel single Core i7 CPU, NVIDIA GeForce GTX-1080 GPU (24 GB memory), and 8 GB RAM. The PC operating system is Ubuntu 16.04.

\subsection{Experimental Results and Discussion} \label{sec:Exp_Dis}

In the experiment, Ms-AFt has been embedded into six detection networks, namely FRCNN\_ResNet101 (FRCNN-A)\citep{HeFRCN2016}, FRCNN\_ResNet101\_receptionV2 (FRCNN-B) \citep{Szegedy2016Inception}, SSD \citep{SSD8026312}, YOLO1 \citep{YOLO1}, YOLO2 \citep{YOLO2}, and R-FCN\_ResNet101 \citep{HeFRCN2016}, to evaluate its detection performance. For a fair comparison, each network uses optimal parameter settings. Fig. \ref{fig:gt} shows the examples of vehicle reference labels selected from the original ground truth and the corresponding detection results on two different datasets. The final quantified result is calculated by the overlap rate of the reference labels and detection results.
\begin{figure*}[!t]
    \begin{center}
        \includegraphics[width=\columnwidth]{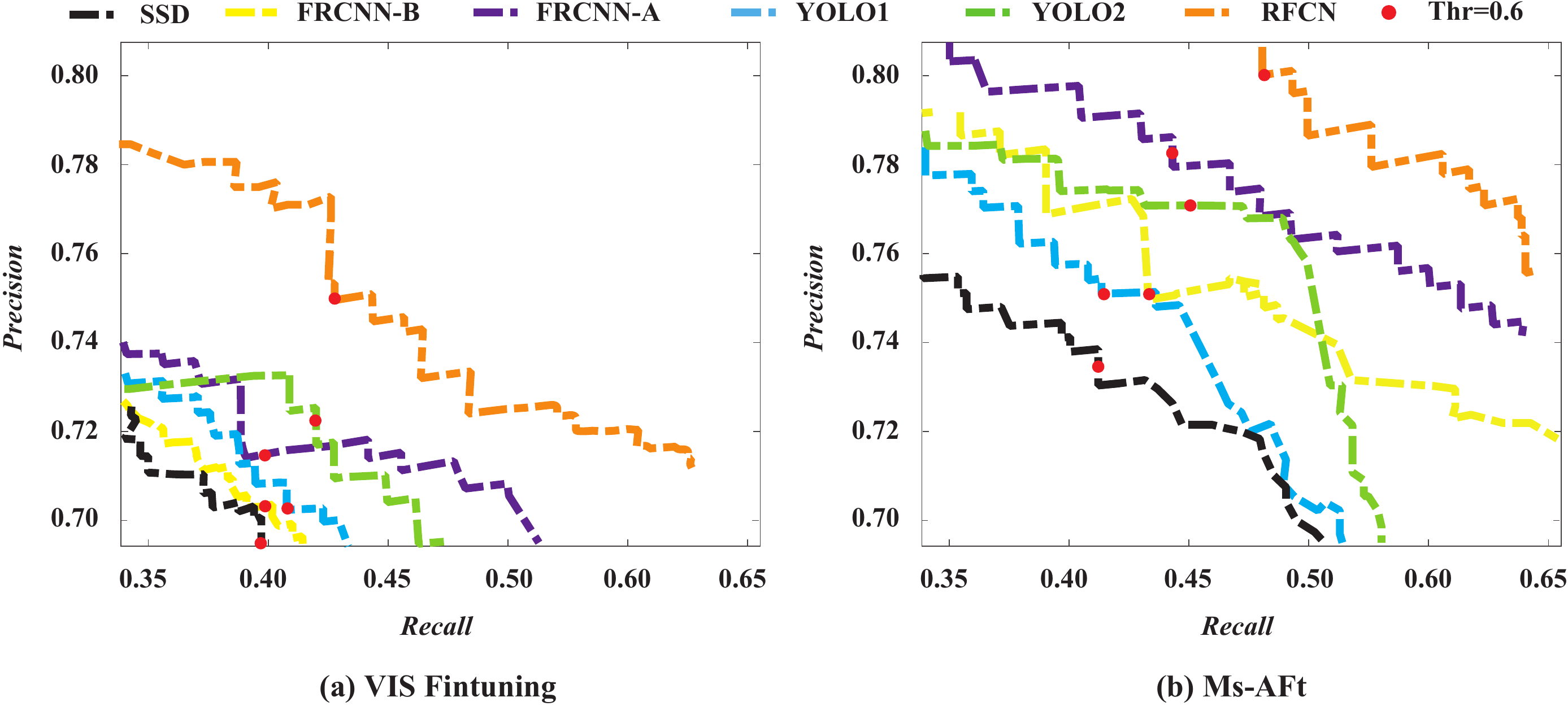}
        \caption{PRCs of the proposed framework on the six example networks for the ISPRS Vaihingen dataset.}
        \label{fig:Vai_PR}
    \end{center}
\end{figure*}
\begin{table*}[!t]
    \centering
    \caption{Performance comparisons of six networks structure for the ISPRS Vaihingen dataset. The best results are highlighted in bold.}\label{tab:Vai}
    \resizebox{\textwidth}{!}{
   \begin{tabular}{c| c c c| c c c| c}
        \toprule
	    & \multirow{2}{*}{Method} & \multirow{2}{*}{Ground Truth} &\multirow{2}{*}{AP} &  \multicolumn{3}{c|}{thr=0.6} & \multirow{2}{*}{Costs/s}   \\
	     &  &   & & Recall & Precision & F1-Score &    \\
	    \hline
		\multirow{6}{*}{VIS Images}& SSD-Ft & 148 & 0.5632 &0.4005 & 0.6925 & 0.5075 & 0.14 \\
		& YOLO1-Ft  & 148 & 0.5927 & 0.4126 & 0.7025 & 0.5199 & 0.13  \\
		& YOLO2-Ft  & 148 & 0.6012 & 0.4258 & 0.7200 & 0.5351 & 0.15  \\
		& FRCNN-A-Ft  & 148 & 0.6000 & 0.4089 & 0.7130 & 0.5198 & 5.54 \\
		& FRCNN-B-Ft  & 148 & 0.5725 & 0.4051 &0.7039 & 0.5142 & 5.98  \\ 
		& R-FCN-Ft & 148 & 0.6500 & 0.4328 & 0.7525 & 0.5495 & 0.32  \\ 
		\hline
		\multirow{6}{*}{Multi-source Images (VIS+DSM)} & SSD-Ms-AFt & 148 & 0.5812 & 0.4221 & 0.7358 & 0.5365 & 0.14 \\
		& YOLO1-Ms-AFt  & 148 & 0.6575 & 0.4225 & 0.7528 & 0.5413 &  \bf 0.13 \\
		& YOLO2-Ms-AFt  & 148 & 0.6600 & 0.4521 & 0.7712 & 0.5700 & 0.15  \\
		& FRCNN-A-Ms-AFt & 148 & 0.6823 & 0.4448 & 0.7821 & 0.5671 & 5.54  \\
		& FRCNN-B-Ms-AFt & 148 & 0.6455 & 0.4341 & 0.7525 & 0.5506& 5.98 \\
		& R-FCN-Ms-AFt & 148 & 0.7277 & \bf 0.4759 & \bf 0.8012 & \bf 0.5971 & 0.32 \\
        \bottomrule
    \end{tabular}}
\end{table*}
\begin{figure*}[h!]
    \begin{center}
        \includegraphics[width=\columnwidth]{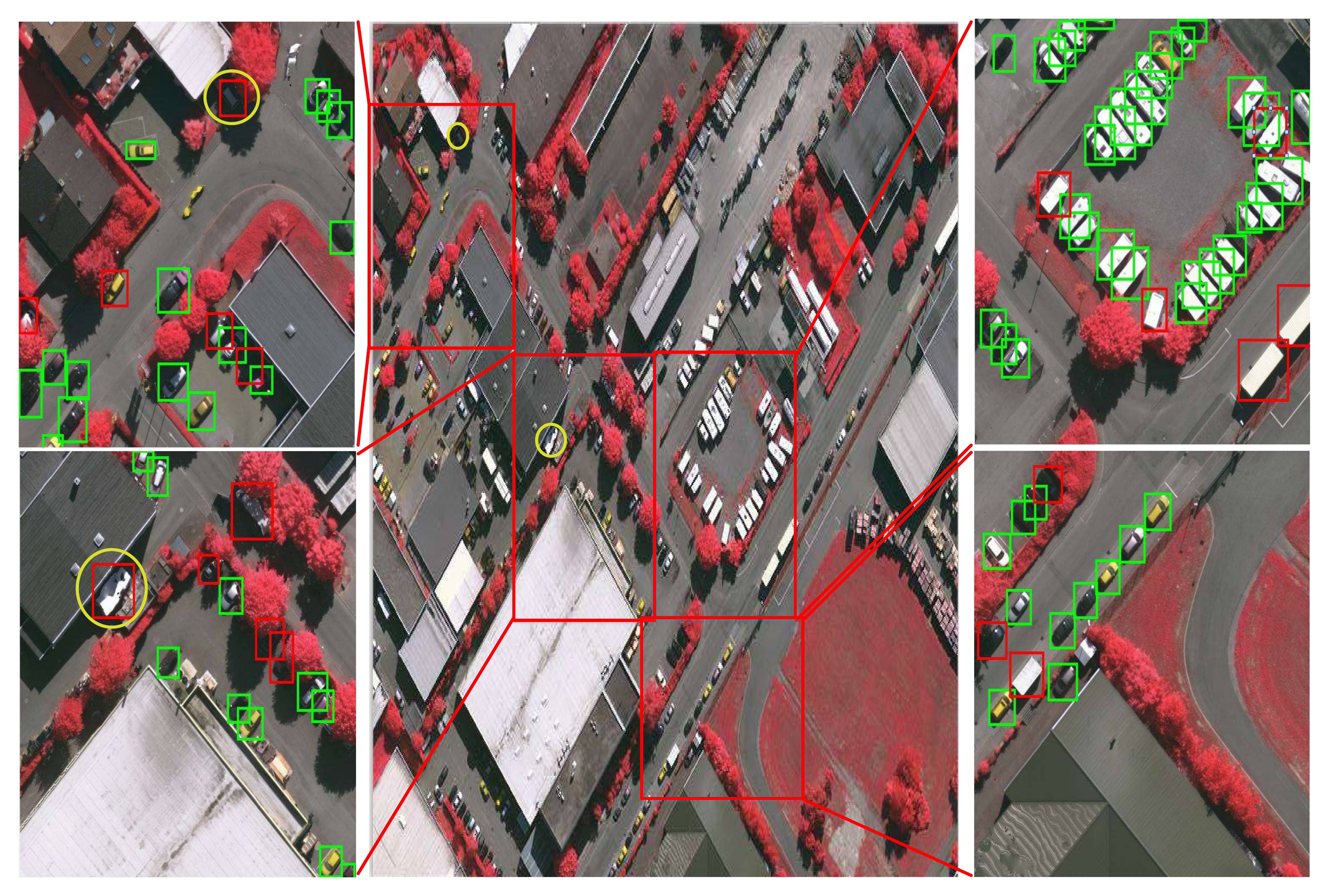}
        \caption{Visual detection results of part areas of the ISPRS Vaihingen dataset. A green box denotes correct detection, a red box denotes leak detection.}
        \label{fig:Vai}
    \end{center}
\end{figure*}

Tables \ref{tab:Vai} - \ref{tab:SAI} provide the quantitative results for three experimental datasets, namely AP and running cost. The objective of this paper is to achieve vehicle auto-labeling by multi-source data while improving its detection performance. So the threshold in PR curves should be chosen to weigh the accuracy and quantity of auto-labeling vehicles, Tables \ref{tab:Vai} - \ref{tab:SAI} also provide the Recall, Precision, and F1-score of all networks when the threshold is 0.6. Overall, the results of the three datasets are similar. One-stage based SSD yields the worst performance. The possible reason is that there is a serious positive and negative sample imbalance problem in the one-stage network. Most of the anchors that the network finally learns are not conducive to the final network learning. While in the two-stage network architecture, e.g., FRCNN, the number of final training anchors is only hundreds or thousands and is useful, which can ensure the network learning optimal to the maximum. YOLO1 is a real-time object detection framework and its detection time is about 0.14s on images with $608\times608$ resolution; however, it is sensitive to the object with a large range of scales and directions. Although YOLO2 improves the scale robustness of the network by introducing the multi-scale feature map, its generalization ability still needs to be improved for small objects. The detection performance of FRCNN with Resnet101 is better than inception\_resnet\_v2\_atrous, our interpretation for this is the atrous convolution could not be reconstructed well in small objects. However, under the same Resnet101 backbone architecture, R-FCN holds a slightly higher precision but lower recall and F1-score than FRCNN. Therefore, for vehicle detection in optical remote sensing images, the two-stage based R-FCN and FRCNN methods are more suitable.
\begin{figure*}[!t]
    \begin{center}
        \includegraphics[width=\columnwidth]{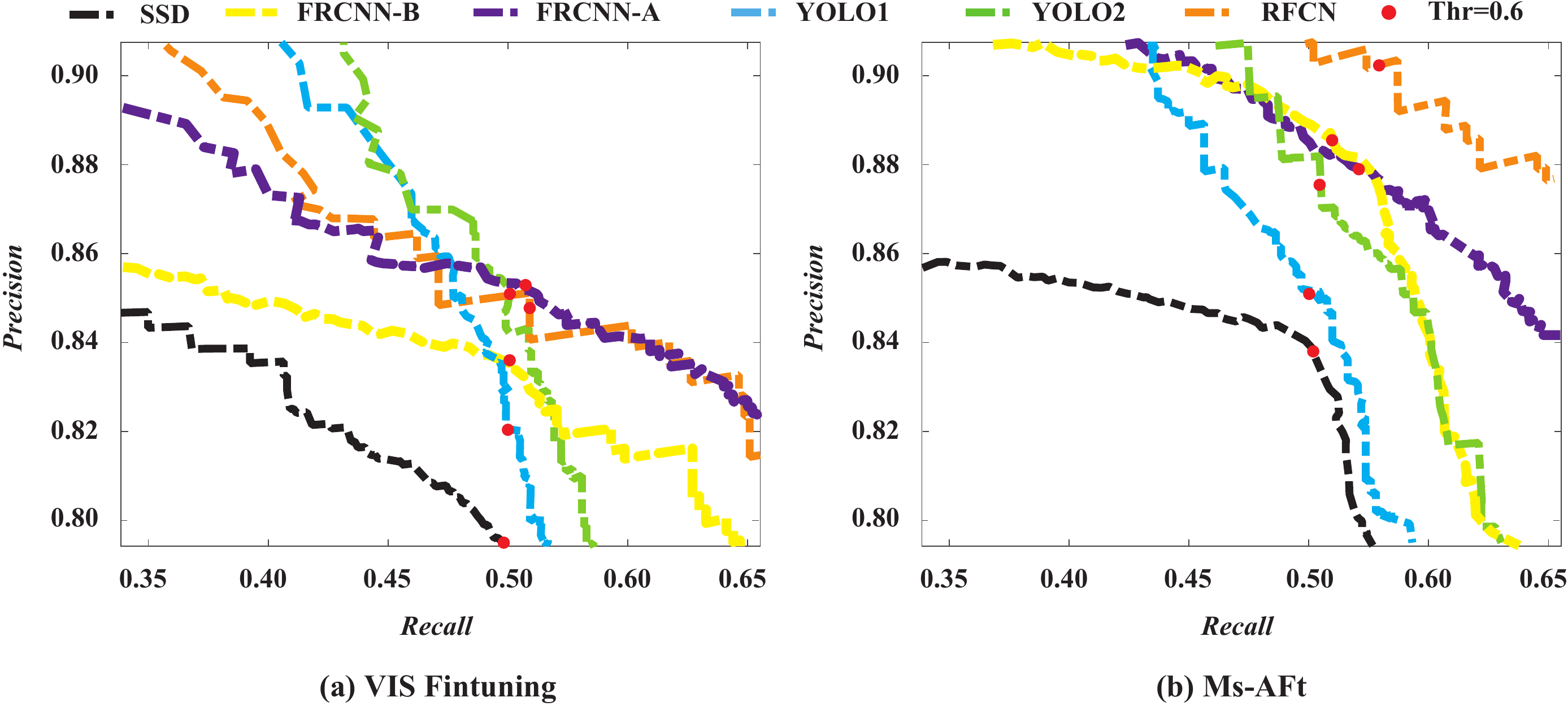}
        \caption{PRCs of the proposed framework on the six example networks for the ISPRS Potsdam dataset.}
        \label{fig:Pos_PR}
    \end{center}
\end{figure*}
\begin{table*}[!t]
    \centering
    \caption{Performance comparisons of six networks structure for the ISPRS Potsdam dataset. The best results are highlighted in bold.\label{tab:Pot}}
    \resizebox{\textwidth}{!}{
   \begin{tabular}{c| c c c| c c c| c}
        \toprule
	     & \multirow{2}{*}{Method} & \multirow{2}{*}{Ground Truth} &\multirow{2}{*}{AP} &  \multicolumn{3}{c|}{thr=0.6} & \multirow{2}{*}{Costs/s}   \\
	     &  &   & & Recall & Precision & F1-Score &   \\ \hline
	    \multirow{6}{*}{VIS Images}& SSD-Ft & 1874 & 0.6458 & 0.4920 & 0.7928 & 0.6072 & 0.15 \\
		& YOLO1-Ft & 1874 & 0.7143 & 0.5008 & 0.8225 & 0.6225 & 0.14 \\
		& YOLO2-Ft & 1874 & 0.7839 & 0.5121 & 0.8488 & 0.6387 & 0.16  \\
		& FRCNN-A-Ft  & 1874 & 0.7498 & 0.5230 & 0.8512 & 0.6479 & 6.34 \\
		& FRCNN-B-Ft  & 1874 & 0.6832 & 0.5091 &0.8376 & 0.6333 & 5.98  \\ 
		& R-FCN-Ft  & 1874 & 0.7577 & 0.5359 & 0.8478 & \textcolor{red}{0.6567} & 0.36  \\ 
		\hline
		\multirow{6}{*}{Multi-source Images (VIS+DSM)} & SSD-Ms-AFt & 1874 & 0.6825 & 0.5214 & 0.8378 & 0.6428 & 0.15 \\
		& YOLO1-Ms-AFt  & 1874 & 0.7705 & 0.5028 & 0.8459 & 0.6307 & \bf 0.14  \\
		& YOLO2-Ms-AFt  & 1874 & 0.8025 & 0.5212 & 0.8788 & 0.6543 & 0.16  \\
		& FRCNN-A-Ms-AFt & 1874 & 0.8100 & 0.5630 & 0.8799 & 0.6866 & 5.98  \\
		& FRCNN-B-Ms-AFt & 1874 & 0.7687 & 0.5155 & 0.8964 & 0.6546 & 6.34 \\
		& R-FCN-Ms-AFt & 1874 & 0.8434 & \bf 0.5779 & \bf 0.9106 & \bf 0.7071 & 0.36 \\
		\bottomrule
    \end{tabular}}
\end{table*}

\begin{figure*}[!t]
    \begin{center}
        \includegraphics[width=\columnwidth]{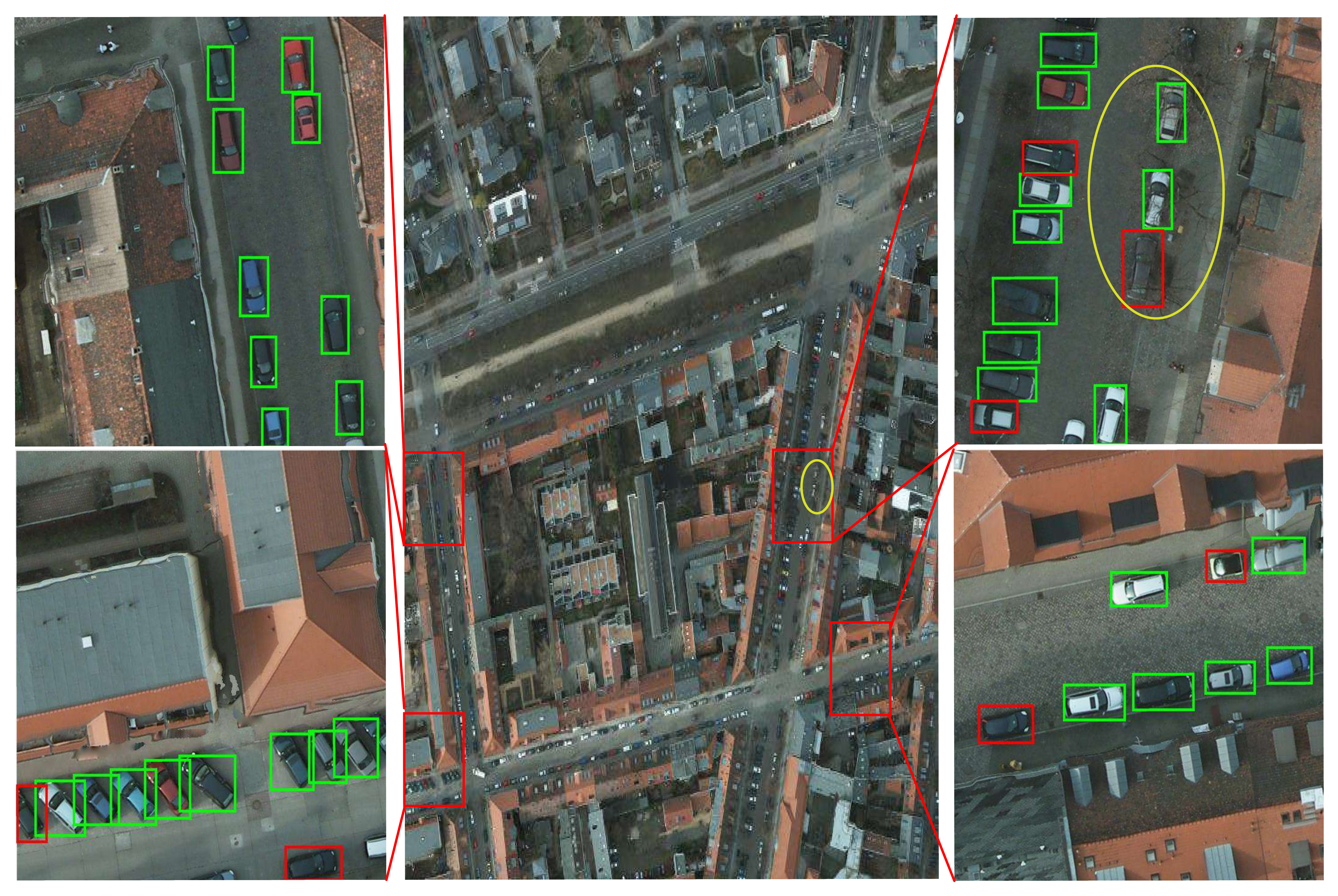}
        \caption{Visual detection results of part areas of ISPRS Potsdam dataset. A green box denotes correct detection, a red box denotes leak detection.}
        \label{fig:Pos}
    \end{center}
\end{figure*}

\textit{1) Vaihingen ISPRS Dataset}: Fig. \ref{fig:Vai} shows the visual detection results of part regions in the ISPRS Vaihingen dataset. Overall, the Vaihingen dataset has the worst detection performance. The possible reason is that this dataset is mainly about complex urban scenes. The dense trees around the building make the parking area have serious occlusion, shadow, etc. These areas constitute an opaque wall, reducing the detection performance of vehicles, especially the black vehicle marked in the yellow circle (e.g., the upper left corner of Fig. \ref{fig:Vai}). Specifically, the number of vehicle training samples is only hundreds. Once there is a certain degree of blurring at the edge of vehicles in DSM, the merged image cannot accurately acquire the part vehicle location, especially for vehicles with serious occlusion. In addition, some vehicles, e.g., the white vehicle is marked in the yellow circle as shown in the lower left corner of Fig. \ref{fig:Vai}, have unknown distortions, and its corresponding merged images inevitably have a fault, which causes the vehicle to be separated into multiple noise-like regions. To some extent, these problems result in a reduction in the number of high quality vehicle samples, making the model under-fitting. Accordingly, we also give the PRCs of VIS fine-tuning and our Ms-AFt networks (see Fig. \ref{fig:Vai_PR}).

\textit{2) Potsdam ISPRS Dataset}: Fig. \ref{fig:Pos} shows the visual detection results of part regions in the ISPRS Potsdam dataset. Overall, the Potsdam dataset offers the highest precision when the threshold is 0.6 than others due to the fact that vehicles in Potsdam have great similarities with the ``small-vehicle'' in the DOTA dataset, thus its auto-labeling results are better in the transfer learning stage than the others. Image resolution in this dataset is higher and only part of the vehicles are covered by tree branches (the white vehicle in the upper right corner of Fig. \ref{fig:Pos}). This enables the merged image to maximize the advantages of each sensor, improving the accuracy of vehicle location and ultimately enhancing the detection performance of the model. The number of vehicles is at least thousands, and the inter-class distance of vehicles is small, which can alleviate the difficulty of vehicle detection to a certain extent. Fig. \ref{fig:Pos_PR} draws the corresponding PRCs with six different networks.

\begin{figure*}[!t]
    \begin{center}
        \includegraphics[width=\columnwidth]{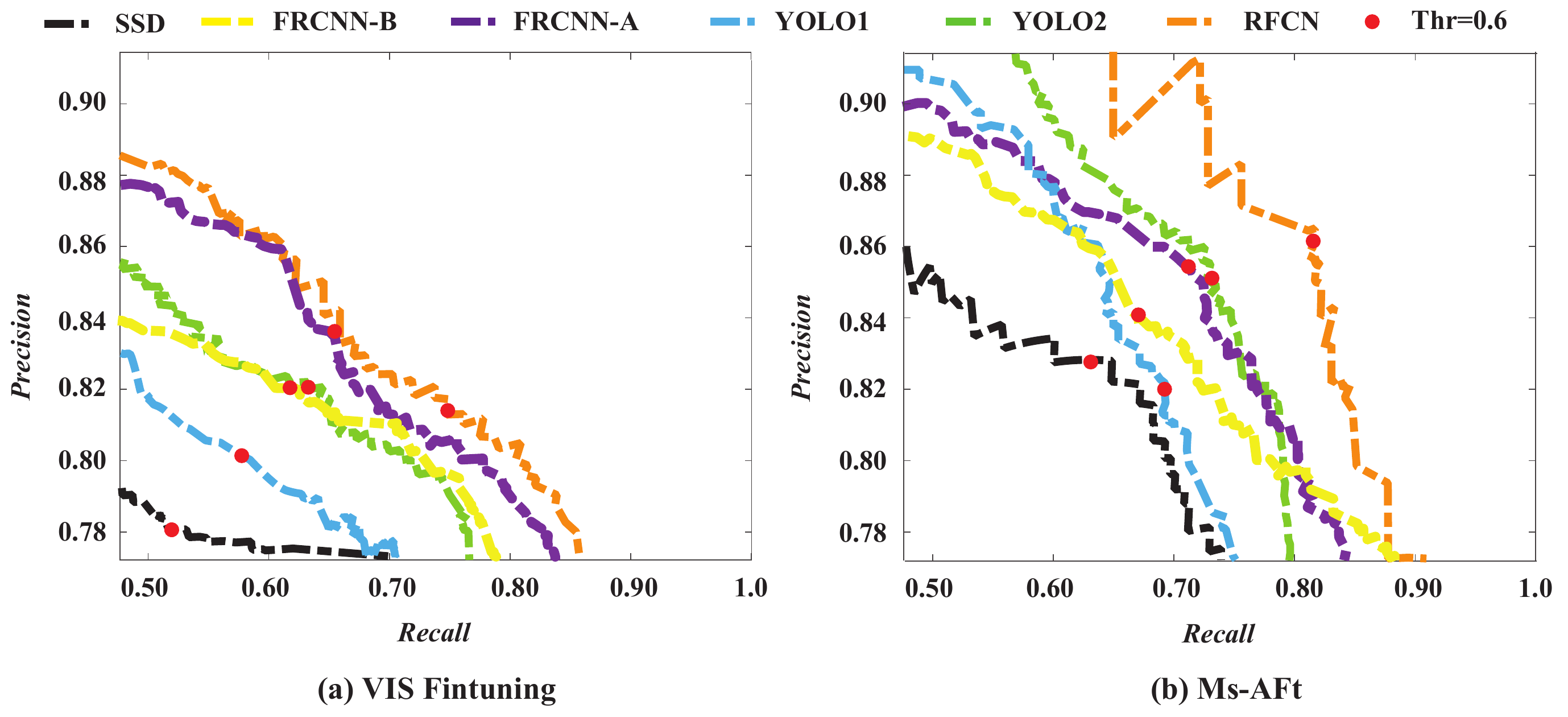}
        \caption{PRCs of the proposed framework on the six example networks for the DLR SAI-LCS dataset.}
        \label{fig:dlr_PR}
    \end{center}
\end{figure*}

\begin{figure*}[!t]
    \begin{center}
        \includegraphics[width=\columnwidth]{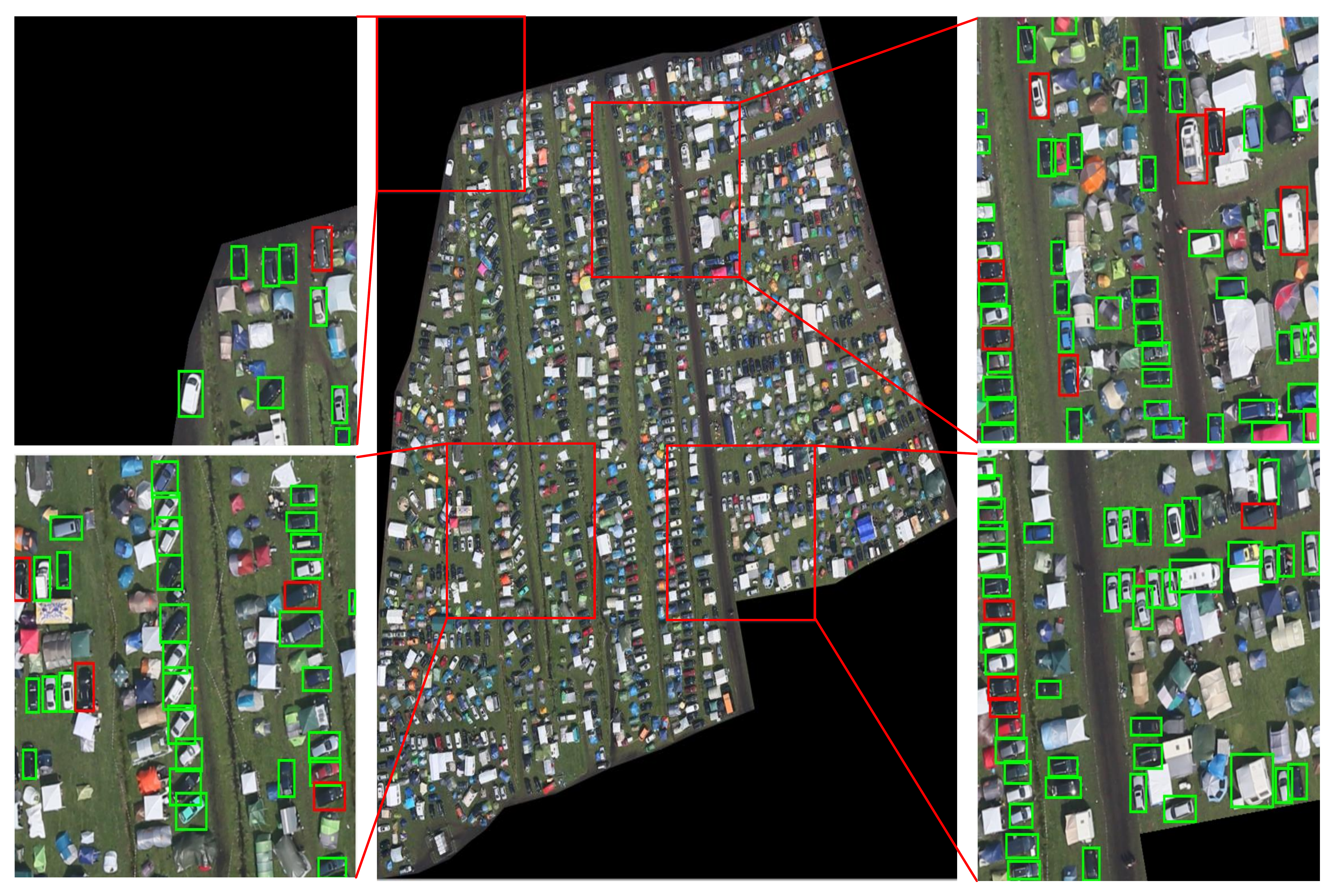}
        \caption{Visual detection results of part areas of the DLR SAI-LCS dataset. A green box denotes correct detection, a red box denotes the leak detection.}
        \label{fig:SAI}
    \end{center}
\end{figure*}

\textit{3) SAI-LCS DLR Dataset}: Fig. \ref{fig:SAI} shows the visual detection results of part regions in the DLR SAI-LCS dataset. Different from the reference DOTA data, the vehicles in the parking area and background of this dataset are more difficult to be detected, which inevitably brings a greater challenge. More specifically, there exists highly cross-mixing or close arrangements between the tents and vehicles in this datasets, while in the DOTA data only vehicles are available. Some residential tents are mistakenly identified as vehicles. For vehicles with low scores, they can be removed by changing the threshold, but those with high scores can be used to optimize training data. There are some false and missing vehicles for pseudo-cars (private car-attached tents), especially white vehicles. The possible reason is that there is a high similarity between this part of the vehicle and the active tent and it is necessary to increase the diversity of vehicles or to refine the vehicles according to its shape. Besides, overcrowded parking can also cause edge blurring, affecting auto-labeling and matching. Similarly, Fig. \ref{fig:dlr_PR} makes a performance comparison between the proposed network and other competitors in the form of PRCs.
\begin{table*}[!t]
    \centering
    \caption{Performance comparisons of six networks structure for the DLR SAI-LCS dataset. The best results are highlighted in bold. \label{tab:SAI}}
    \resizebox{\textwidth}{!}{
   \begin{tabular}{c| c c c| c c c| c}
        \toprule
	    & \multirow{2}{*}{Method} & \multirow{2}{*}{Ground Truth} &\multirow{2}{*}{AP} &  \multicolumn{3}{c|}{thr=0.6} & \multirow{2}{*}{Costs/s}   \\
	     &  &   & & Recall & Precision & F1-Score &  \\ \hline
		\multirow{6}{*}{VIS Images}& SSD-Ft & 2313 & 0.7334 & 0.5162 & 0.7804 & 0.6214 & 0.18 \\
		& YOLO1-Ft  & 2313 & 0.7515 & 0.5879 & 0.8007 & 0.6780 & 0.16  \\
		& YOLO2-Ft  & 2313 & 0.7809 & 0.6423 & 0.8221 & 0.7212 & 0.18  \\
		& FRCNN-A-Ft& 2313 & 0.8000 & 0.6597 & 0.8380 & 0.7383 & 6.36  \\
		& FRCNN-B-Ft & 2313 & 0.7600 & 0.6342 &0.8225 & 0.7162 & 6.68  \\ 
		& R-FCN-Ft  & 2313 & 0.8065 & 0.7518 & 0.8164 & 0.7828 & 0.40  \\ 
		\hline
		\multirow{6}{*}{Multi-source Images (VIS+DSM)} & SSD-Ms-AFt & 2313 & 0.7803 & 0.6373 & 0.8385 & 0.7241 & 0.18 \\
		& YOLO1-Ms-AFt  & 2313 & 0.7900 & 0.7012 & 0.8225 & 0.7570 & \bf 0.16  \\
		& YOLO2-Ms-AFt & 2313 & 0.8354 & 0.7423 & 0.8575 & 0.7985 & 0.18  \\
		& FRCNN-A-Ms-AFt & 2313 & 0.8312 & 0.7300 & 0.8601 & 0.7871 & 6.36  \\
		& FRCNN-B-Ms-AFt & 2313 & 0.7978 & 0.7013 & 0.8422 & 0.7653& 6.68 \\
		& R-FCN-Ms-AFt & 2313 & 0.8525 & \bf 0.8313 & \bf 0.8612 & \bf 0.8463 & 0.40 \\
		\bottomrule
    \end{tabular}}
\end{table*}
		
\subsection{Ablation Analysis}

There are three branches in our framework, namely the fine-tuning, object segmentation and active attention branches. As a result, we perform the ablation analysis on the three datasets to investigate the effectiveness of Ms-AFt. Table \ref{tab:branch} takes the R-FCN network as an example, and lists the performance gain by integrating three branches under this network. In view of the contribution of multi-source data to the vehicles auto-labeling in the object segmentation branch, and even for the proposed framework, we list the vehicle detection performance in case of multi-source data and without DSM. It can be seen that multi-source data results are much better than those without DSM. The possible reason is that single VIS images cannot effectively separate and label vehicles with occluded, background clutter, illumination, shadow, especially extreme adhesion. However, it should be noted that Segmentation \& Attention can locate vehicles in unlabeled images but is time consuming. Therefore, the performance of Segmentation \& Attention on the three datasets is relatively inferior to that of the fine-tuning branch due to insufficient high quality vehicle training samples. Finally, the proposed Ms-AFt outperforms single branch detection dramatically. Across three datasets, Vaihingen in terms of vehicles severely occluded results in only 0.05 F1-score improvement. For Potsdam, enough vehicle samples make the precision have 0.2 improvements, and incomplete vehicle types make the recall have only 0.03 improvements. For SAI-LCS, there is an obvious improvement in recall, owing to the diversity of vehicles.
\begin{table*}[!t]
    \centering
    \footnotesize
    \caption{Performance comparisons of ablation studies for three datasets. The best results are highlighted in bold.}
    \resizebox{\textwidth}{!}{
   \begin{tabular}{c c c c c c}
        \toprule 
		Dataset & DataSource & Branch & Recall & Precion & F1-Score\\ \hline 
		\multirow{5}{*}{ISPRS\_Vaihingen} & VIS& Fine-tuning & 0.4328 & 0.7525 & 0.5495\\ 
		\cline{2-6} & VIS & Segmentation \& Attention & 0.2000 & 0.5034 & 0.2863  \\ 
		& VIS & VIS-AFt & 0.3434 &0.6489 & 0.4491 \\ 
		\cline{2-6} & VIS+DSM & Segmentation \& Attention & 0.3214 & 0.6128 & \textcolor{red}{0.4217}  \\ 
		& VIS+DSM &  Ms-AFt & \bf 0.4759 & \bf 0.8012 & \bf 0.5971 \\
		\hline \hline
		\multirow{5}{*}{ISPRS\_Potsdam} &  VIS& Fine-tuning & 0.5559 & 0.7378 & \textcolor{red}{0.6341}\\ 
		\cline{2-6} & VIS & Segmentation \& Attention & 0.3170 & 0.5343 & \textcolor{red}{0.3979}  \\ 
		& VIS &  VIS-AFt & 0.4078 & 0.7349 &0.5245 \\
		\cline{2-6} & VIS+DSM & Segmentation \& Attention & 0.4225 & 0.6170 & 0.5016  \\ 
		& VIS+DSM &  Ms-AFt & \bf 0.5779 & \bf 0.9106 & \bf 0.7071 \\
		\hline \hline
		\multirow{5}{*}{DLR\_SAI-LCS} & VIS & Fine-tuning & 0.5260 & 0.8429 & 0.6477\\ 
		\cline{2-6} & VIS & Segmentation \& Attention & 0.3388 & 0.4000 & 0.3669  \\ 
		& VIS  & VIS-AFt & 0.5565 & 0.6109 & 0.5824 \\
		\cline{2-6} & VIS+DSM & Segmentation \& Attention & 0.4695 & 0.5480 & 0.5057  \\ 
		& VIS+DSM  & Ms-AFt & \bf 0.8313 & \bf 0.8612 & \bf 0.8463 \\
		\bottomrule
    \end{tabular}}
    \label{tab:branch}
\end{table*}

\subsection{Influence of Resolution}

\begin{figure*}[!t]
	  \begin{center}
	  	\subfigure[ISPRS\_Vaihingen Dataset]{
			\includegraphics[width=0.45\textwidth]{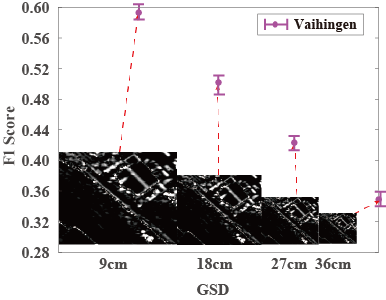}
		}
		\subfigure[ISPRS\_Potsdam Dataset]{
			\includegraphics[width=0.45\textwidth]{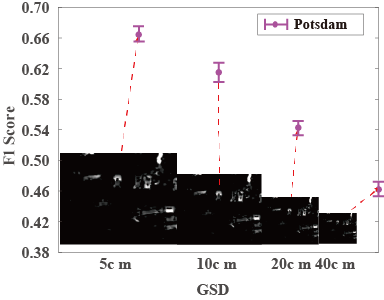}
		}
		\subfigure[DLR SAI-LSC Dataset]{
			\includegraphics[width=0.45\textwidth]{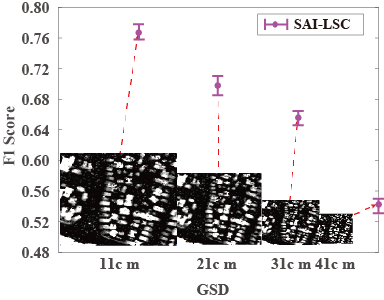}
		}
        \caption{Image resolutions analysis of the proposed AFts in respect to three competitive datasets. Detection performance analysis of active fine-tuning networks with different resolutions of three datasets}
        \label{fig:resolution}
        \end{center}
\end{figure*}

We experimentally analyze and discuss the potential influences under the different resolutions of the DSM images. The original image is sequentially downsampled and upsampled, which reduces the image resolution while adapting to the input of the detection network. Fig. \ref{fig:resolution} shows the detection performance with different GSD resolutions. Overall, the F1-score is significantly reduced when the GSD in three datasets drops to three times the initial GSD, with a maximum loss of about 20\%. Specifically, the F1-score of the Vaihingen dataset loss the most, the resolution of GSD is reduced by half each time, and the performance is lost by more than 10\%. This is due to the vehicles in vision images having serious occlusion problems (including shadows and trees), and some vehicles are densely arranged. The resolution reduction of GSD not only reduces shadowed vehicle positioning but also the fuzzy elevation information is not conducive to assist densely arranged vehicle positioning.
For Potsdam in terms of more dark vehicles, their reduced detachability when GSD resolution decreased. Compared with the Vaihingen dataset, vehicle detection in Potsdam has less performance degradation under the sparse arrangement in a simple background. The SAI-LCS image contains two object types, namely vehicles and tents. They are similar in shape and densely arranged. In the absence of a large number of reference labeled complex vehicles, e.g., RVs and campers, the resolution reduction of GSD have seriously affected the refinement of ground object areas and the screening of high quality vehicles. Fortunately, there is no high building and lighting shadows occlusion in this dataset; vehicle detection performance loss is about 8\% when the resolution of GSD is reduced by half.

\section{Conclusion}\label{sec:Con}

Considering the complexity and inconsistency of manual labeling for objects with complex changes in airborne optical remote sensing images, a vehicle auto-labeling and detection framework of multi-source data using active fine-tuning network (Ms-AFt) was developed. Based on multi-source data, the proposed method attempted to automatically label and detect vehicles using a labeled but task-independent reference dataset. 

The experiments employed HBB-based labeling to verify the effectiveness of Ms-AFt in two public datasets and one non-public dataset. Considering HBB is sensitive to objects with direction variation, we will focus on the oriented bounding boxes (OBB) based labeling by designing direction-insensitive or direction-robust auto-labeling method, further resolving severe adhesions between objects and improving the label quality of vehicles or other direction variation objects in future work. Besides, we will investigate other state-of-the-art networks, e.g., YOLO3, and lightweight networks, e.g., shufflenet V2, GhostNet, in balancing accuracy and speed requirements of object detection in practical applications. Also, we will integrate contextual semantic information and multi-class object detection to further distinguish between different types of vehicles, such as private cars, transport vehicles, motor homes, and camping trailers.

\section*{Acknowledgement}

This work was supported, in part by the National Natural Science Foundation of China under Grant 61922013, 61421001, and U1833203, and partly by the Fundamental Research Funds
for the Central Universities under Grant 3052019116.



\bibliographystyle{elsarticle-harv}
\bibliography{mybibfile}

\begin{thebibliography}{53}
\expandafter\ifx\csname natexlab\endcsname\relax\def\natexlab#1{#1}\fi
\expandafter\ifx\csname url\endcsname\relax
  \def\url#1{\texttt{#1}}\fi
\expandafter\ifx\csname urlprefix\endcsname\relax\def\urlprefix{URL }\fi

\bibitem[{Achanta et~al.(2012)Achanta, Shaji, Smith, Lucchi, Fua, and
  S{\"u}sstrunk}]{achanta2012slic}
Achanta, R., Shaji, A., Smith, K., Lucchi, A., Fua, P., S{\"u}sstrunk, S.,
  2012. Slic superpixels compared to state-of-the-art superpixel methods. IEEE
  Trans. Pattern Anal. Mach. Intell 34~(11), 2274--2282.

\bibitem[{Aldeborgh et~al.(2017)Aldeborgh, Ouzounis, and Stamatiou}]{inp}
Aldeborgh, N., Ouzounis, G., Stamatiou, K., 2017. Unsupervised object detection
  on remote sensing imagery using hierarchical image representations and deep
  learning. In: Proc. Int. Conf. IBig Data from Space (BiDS). pp. 255--258.

\bibitem[{Arivalagan and Venkatachalapathy(2015)}]{Arivalagan2015Vehicle}
Arivalagan, S., Venkatachalapathy, K., 2015. Vehicle detection in traffic
  videos using differential evolution algorithm trained neural network. Int. J.
  Applied Engineering Research 10~(6), 14691--14702.

\bibitem[{Audebert et~al.(2017)Audebert, Bertrand, and
  Sébastien}]{Nicolas2017Segment}
Audebert, N., Bertrand, L., Sébastien, L., 2017. Segment-before-detect:
  Vehicle detection and classification through semantic segmentation of aerial
  images. Remote sens. 9~(4), 368.

\bibitem[{Cao et~al.(2016)Cao, Luo, Chen, Sheng, Wang, Wang, and
  Ji}]{Cao2016Weakly}
Cao, L., Luo, F., Chen, L., Sheng, Y., Wang, H., Wang, C., Ji, R., 2016. Weakly
  supervised vehicle detection in satellite images via multi-instance
  discriminative learning. Pattern Recogn. 64, 417--424.

\bibitem[{Chai(2016)}]{Chai2016A}
Chai, D., 2016. A probabilistic framework for building extraction from airborne
  color image and dsm. IEEE J. Sel. Topics Appl. Earth Observ. Remote Sens.
  10~(3), 948--959.

\bibitem[{Chen et~al.(2018{\natexlab{a}})Chen, Liu, Hu, and Pan}]{8519132}
Chen, G., Liu, L., Hu, W., Pan, Z., 2018{\natexlab{a}}. Semi-supervised object
  detection in remote sensing images using generative adversarial networks. In:
  Proc. IEEE. Int. Conf. Geoscience and Remote Sensing Symposium (IGARSS). pp.
  2503--2506.

\bibitem[{Chen et~al.(2018{\natexlab{b}})Chen, Zhang, and Ouyang}]{Chen}
Chen, Z., Zhang, T., Ouyang, C., 2018{\natexlab{b}}. End-to-end airplane
  detection using transfer learning in remote sensing images. Remote sens.
  10~(1), 139.

\bibitem[{Cheng and Han(2016)}]{cheng2016survey}
Cheng, G., Han, J., 2016. A survey on object detection in optical remote
  sensing images. ISPRS J. Photogramm. Remote Sens. 117, 11--28.

\bibitem[{Cheng et~al.(2014)Cheng, Han, Zhou, and Guo}]{Cheng2014}
Cheng, G., Han, J., Zhou, P., Guo, L., 2014. Multi-class geospatial object
  detection and geographic image classification based on collection of part
  detectors. ISPRS J. Photogramm. Remote Sens. 98~(1), 119--132.

\bibitem[{Cheng et~al.(2018)Cheng, Han, Zhou, and Xu}]{cheng2018learning}
Cheng, G., Han, J., Zhou, P., Xu, D., 2018. Learning rotation-invariant and
  fisher discriminative convolutional neural networks for object detection.
  IEEE Trans. on Image Processing 28~(1), 265--278.

\bibitem[{Cheng et~al.(2020)Cheng, Si, Hong, Yao, and Guo}]{cheng2020cross}
Cheng, G., Si, Y., Hong, H., Yao, X., Guo, L., 2020. Cross-scale feature fusion
  for object detection in optical remote sensing images. IEEE Geosci. Remote
  Sens. Lett.

\bibitem[{Cheng et~al.(2016)Cheng, Zhou, and Han}]{Cheng2016}
Cheng, G., Zhou, P., Han, J., 2016. Learning rotation-invariant convolutional
  neural networks for object detection in vhr optical remote sensing images.
  IEEE Trans. Geosci. Remote Sens. 54~(12), 7405--7415.

\bibitem[{Dai et~al.(2016)Dai, Li, He, and Sun}]{HeFRCN2016}
Dai, J., Li, Y., He, K., Sun, J., 2016. R-fcn: object detection via
  region-based fully convolutional networks. In: Proc. IEEE Int. Conf. on
  Computer Vision and Pattern Recognition (CVPR). pp. 379--387.

\bibitem[{Ester et~al.(1996)Ester, Kriegel, Sander, and Xu}]{ester1996density}
Ester, M., Kriegel, H., Sander, J., Xu, X., 1996. A density-based algorithm for
  discovering clusters in large spatial databases with noise. In: Proc. KDD.
  Vol.~96. pp. 226--231.

\bibitem[{Gintautas et~al.(2009)Gintautas, Franz, and
  Peter}]{Palubinskas2009Detection}
Gintautas, P., Franz, K., Peter, R., 2009. Detection of traffic congestion in
  optical remote sensing imagery. In: Proc. IEEE. Int. Conf. Geoscience and
  Remote Sensing Symposium (IGARSS). Vol.~2. pp. II--426.

\bibitem[{Gstaiger et~al.(2018)Gstaiger, Tian, Kiefl, and Kurz.}]{Tian2018}
Gstaiger, V., Tian, J., Kiefl, R., Kurz., F., 2018. 2d vs. 3d change detection
  using aerial imagery to support crisis management of large-scale events.
  Remote Sens. 10~(12), 2054.

\bibitem[{Han et~al.(2015)Han, Zhang, Cheng, Guo, and Ren}]{6991537}
Han, J., Zhang, D., Cheng, G., Guo, L., Ren, J., 2015. Object detection in
  optical remote sensing images based on weakly supervised learning and
  high-level feature learning. IEEE Trans. Geosci. Remote Sens. 53~(6),
  3325--3337.

\bibitem[{He et~al.(2016)He, Zhang, Ren, and Sun}]{He2015ResNet}
He, K., Zhang, X., Ren, S., Sun, J., 2016. Deep residual learning for image
  recognition. In: Proceedings of the IEEE conference on computer vision and
  pattern recognition. pp. 770--778.

\bibitem[{He et~al.(2019)He, Wang, Ghamisi, Li, and Chen}]{8480863}
He, X., Wang, A., Ghamisi, P., Li, G., Chen, Y., 2019. Lidar data
  classification using spatial transformation and cnn. IEEE Geosci. Remote
  Sens. Lett. 16~(1), 125--129.

\bibitem[{Hendrik et~al.(2018)Hendrik, Dimitri, and Wolfgang}]{Hen2018}
Hendrik, S., Dimitri, B., Wolfgang, M., 2018. Object-based detection of
  vehicles using combined optical and elevation data. ISPRS J. Photogramm.
  Remote Sens. 136, 85--105.

\bibitem[{Hong et~al.(2020)Hong, Wu, Ghamisi, Chanussot, Yokoya, and
  Zhu}]{hong2020invariant}
Hong, D., Wu, X., Ghamisi, P., Chanussot, J., Yokoya, N., Zhu, X.~X., 2020.
  Invariant attribute profiles: A spatial-frequency joint feature extractor for
  hyperspectral image classification. IEEE Trans. Geosci. Remote Sens. 58~(6),
  3791--3808.

\bibitem[{Hong et~al.(2019{\natexlab{a}})Hong, Yokoya, Chanussot, and
  Zhu}]{hong2018augmented}
Hong, D., Yokoya, N., Chanussot, J., Zhu, X.~X., 2019{\natexlab{a}}. An
  augmented linear mixing model to address spectral variability for
  hyperspectral unmixing. IEEE Trans. Image Process. 28~(4), 1923--1938.

\bibitem[{Hong et~al.(2019{\natexlab{b}})Hong, Yokoya, Chanussot, and
  Zhu}]{hong2019cospace}
Hong, D., Yokoya, N., Chanussot, J., Zhu, X.~X., 2019{\natexlab{b}}. Co{S}pace:
  Common subspace learning from hyperspectral-multispectral correspondences.
  IEEE Trans. Geosci. Remote Sens. 57~(7), 4349--4359.

\bibitem[{Hong et~al.(2019{\natexlab{c}})Hong, Yokoya, Ge, Chanussot, and
  Zhu}]{hong2019learnable}
Hong, D., Yokoya, N., Ge, N., Chanussot, J., Zhu, X., 2019{\natexlab{c}}.
  Learnable manifold alignment ({L}e{MA}): A semi-supervised cross-modality
  learning framework for land cover and land use classification. ISPRS J.
  Photogramm. Remote Sens. 147, 193--205.

\bibitem[{Huang et~al.(2020)Huang, Hong, Xu, Yao, and Stilla}]{huang2019multi}
Huang, R., Hong, D., Xu, Y., Yao, W., Stilla, U., 2020. Multi-scale local
  context embedding for lidar point cloud classification. IEEE Geosci. Remote
  Sens. Lett. 17~(4), 721--725.

\bibitem[{Ji et~al.(2019)Ji, Gao, Mei, and Li}]{8694857}
Ji, H., Gao, Z., Mei, T., Li, Y., 2019. Improved faster r-cnn with multiscale
  feature fusion and homography augmentation for vehicle detection in remote
  sensing images. IEEE Geosci. Remote Sens. Lett. 16~(11), 1761--1765.

\bibitem[{Joseph and Ali(2017)}]{YOLO2}
Joseph, R., Ali, F., 2017. Yolo9000: Better, faster, stronger. In: Proc. IEEE
  Int. Conf. on Computer Vision and Pattern Recognition (CVPR). pp. 6517--6525.

\bibitem[{Joseph et~al.(2016)Joseph, Santosh, Ross, and Ali}]{YOLO1}
Joseph, R., Santosh, D., Ross, G., Ali, F., 2016. You only look once: Unified,
  real-time object detection. In: Proc. IEEE Int. Conf. on Computer Vision and
  Pattern Recognition (CVPR). p. 779–788.

\bibitem[{Kang et~al.(2020)Kang, Hong, Liu, Baier, Yokoya, and
  Demir}]{kang2020learning}
Kang, J., Hong, D., Liu, J., Baier, G., Yokoya, N., Demir, B., 2020. Learning
  convolutional sparse coding on complex domain for interferometric phase
  restoration. IEEE Trans. Neural Netw. Learn. Syst.DOI:
  10.1109/TNNLS.2020.2979546.

\bibitem[{Kang and Gellert(2015)}]{Liu2015Fast}
Kang, L., Gellert, M., 2015. Fast multiclass vehicle detection on aerial
  images. IEEE Geosci. Remote Sens. Lett. 12~(9), 1938--1942.

\bibitem[{Karim et~al.(2014)Karim, Moutakki, Ayaou, and
  Amghar}]{Karim2014Prototype}
Karim, A., Moutakki, Z., Ayaou, T., Amghar, A., 2014. Prototype of an embedded
  system using stratix iii fpga for vehicle detection and traffic management.
  In: Proc. Int. Conf. Multimedia Computing and Systems (ICMCS). pp. 141--146.

\bibitem[{Li et~al.(2020)Li, Wan, Cheng, Meng, and Han}]{li2020object}
Li, K., Wan, G., Cheng, G., Meng, L., Han, J., 2020. Object detection in
  optical remote sensing images: A survey and a new benchmark. ISPRS J.
  Photogramm. Remote Sens. 159, 296--307.

\bibitem[{Lin et~al.(2016)Lin, Fu, Wang, Xu, and Sun}]{Lin2016MARTA}
Lin, D., Fu, K., Wang, Y., Xu, G., Sun, X., 2016. Marta gans: Unsupervised
  representation learning for remote sensing image classification. IEEE Geosci.
  Remote Sens. Lett. 14~(11), 2092--2096.

\bibitem[{Mandal et~al.(2019)Mandal, Shah, Meena, Devi, and
  Vipparthi}]{8755462}
Mandal, M., Shah, M., Meena, P., Devi, S., Vipparthi, S.~K., 2019. Avdnet: A
  small-sized vehicle detection network for aerial visual data. IEEE Geosci.
  Remote Sens. Lett. 17~(3), 494--498.

\bibitem[{Marmanis et~al.(2016)Marmanis, Schindler, Wegner, Galliani, Datcu,
  and Stilla}]{Marmanis2016}
Marmanis, D., Schindler, K., Wegner, J., Galliani, S., Datcu, M., Stilla, U.,
  2016. Classification with an edge: Improving semantic image segmentation with
  boundary detection. ISPRS J. Photogramm. Remote Sens. 135, 158--172.

\bibitem[{Niessner et~al.(2017)Niessner, Schilling, and Jutzi}]{Nie2017}
Niessner, R., Schilling, H., Jutzi, B., 2017. Investigations on the potential
  of convolutional neural networks for vehicle classification based on rgb and
  lidar data. ISPRS Annals of the Photogrammetry, Remote Sensing and Spatial
  Information Sciences 4, 115.

\bibitem[{Ning et~al.(2017)Ning, Zhou, Song, and Tang}]{SSD8026312}
Ning, C., Zhou, H., Song, Y., Tang, J., 2017. Inception single shot multibox
  detector for object detection. In: Proc. IEEE Int. Conf. on Multimedia Expo
  Workshops (ICMEW). pp. 549--554.

\bibitem[{Pang et~al.(2017)Pang, Sun, Ren, Yang, and Yan}]{Pang2018Cascade}
Pang, J., Sun, W., Ren, J.~S., Yang, C., Yan, Q., 2017. Cascade residual
  learning: A two-stage convolutional neural network for stereo matching. In:
  Proc. IEEE Int. Conf. on Computer Vision Workshops. pp. 887--895.

\bibitem[{Rasti et~al.(2020)Rasti, Hong, Hang, Ghamisi, Kang, Chanussot, and
  Benediktsson}]{rasti2020feature}
Rasti, B., Hong, D., Hang, R., Ghamisi, P., Kang, X., Chanussot, J.,
  Benediktsson, J., 2020. Feature extraction for hyperspectral imagery: The
  evolution from shallow to deep (overview and toolbox). IEEE Geosci. Remote
  Sens. Mag.DOI: 10.1109/MGRS.2020.2979764.

\bibitem[{Schilling et~al.(2018)Schilling, Bulatov, Niessner, Middelmann, and
  Soergel}]{Schilling2018Detection}
Schilling, H., Bulatov, D., Niessner, R., Middelmann, W., Soergel, U., 2018.
  Detection of vehicles in multisensor data via multibranch convolutional
  neural networks. IEEE J. Sel. Topics Appl. Earth Observ. Remote Sens.
  11~(11), 4299--4316.

\bibitem[{Simonyan and Zisserman(2014)}]{Karen2014VGG}
Simonyan, K., Zisserman, A., 2014. Very deep convolutional networks for
  large-scale image recognition. arXiv preprint arXiv:1409.1556.

\bibitem[{{Sumbul} et~al.(2019){Sumbul}, {Cinbis}, and {Aksoy}}]{8648481}
{Sumbul}, G., {Cinbis}, R.~G., {Aksoy}, S., 2019. Multisource region attention
  network for fine-grained object recognition in remote sensing imagery. IEEE
  Trans. Geosci. Remote Sens. 57~(7), 4929--4937.

\bibitem[{Szegedy et~al.(2017)Szegedy, Ioffe, Vanhoucke, and
  Alemi}]{Szegedy2016Inception}
Szegedy, C., Ioffe, S., Vanhoucke, V., Alemi, A.~A., 2017. Inception-v4,
  inception-resnet and the impact of residual connections on learning. In:
  Proc. Thirty-first AAAI conference on artificial intelligence (AAAI).

\bibitem[{Wei et~al.(2013)Wei, Zhou, and Zheng}]{6723710}
Wei, H., Zhou, G., Zheng, Z., 2013. Detection of traffic congestion in optical
  remote sensing imagery. In: Proc. IEEE. Int. Conf. Geoscience and Remote
  Sensing Symposium (IGARSS). pp. 4002--4005.

\bibitem[{Wen et~al.(2019)Wen, Bruno, Konrad, and Nicola}]{Wen2019}
Wen, X., Bruno, V., Konrad, S., Nicola, P., 2019. Street-side vehicle
  detection, classification and change detection using mobile laser scanning
  data. ISPRS J. Photogramm. Remote Sens. 114, 166--178.

\bibitem[{Weng et~al.(2014)Weng, Fu, and Gao}]{weng2014generating}
Weng, Q., Fu, P., Gao, F., 2014. Generating daily land surface temperature at
  landsat resolution by fusing landsat and modis data. Remote Sens. Environ.
  145, 55--67.

\bibitem[{Weng et~al.(2018)Weng, Quattrochi, and Gamba}]{weng2018urban}
Weng, Q., Quattrochi, D., Gamba, P.~E., 2018. Urban remote sensing. CRC press.

\bibitem[{Wu et~al.(2020)Wu, Hong, Chanussot, Xu, Tao, and
  Wang}]{wu2019fourier}
Wu, X., Hong, D., Chanussot, J., Xu, Y., Tao, R., Wang, Y., 2020. Fourier-based
  rotation-invariant feature boosting: An efficient framework for geospatial
  object detection. IEEE Geosci. Remote Sens. Lett. 17~(2), 302--306.

\bibitem[{Wu et~al.(2019)Wu, Hong, Tian, Chanussot, Li, and Tao}]{Wu2018}
Wu, X., Hong, D., Tian, J., Chanussot, J., Li, W., Tao, R., 2019. Orsim
  detector: A novel object detection framework in optical remote sensing
  imagery using spatial-frequency channel features. IEEE Trans. Geosci. Remote
  Sens. 57~(7), 5146--5158.

\bibitem[{Xia et~al.(2018)Xia, Bai, Ding, Zhu, Belongie, Luo, Datcu, Pelillo,
  and Zhang}]{xia2018dota}
Xia, G., Bai, X., Ding, J., Zhu, Z., Belongie, S., Luo, J., Datcu, M., Pelillo,
  M., Zhang, L., 2018. Dota: A large-scale dataset for object detection in
  aerial images. In: Proc. IEEE Int. Conf. on Computer Vision and Pattern
  Recognition (CVPR). pp. 3974--3983.

\bibitem[{Yang et~al.(2018)Yang, Li, and Lin}]{Yang2018Vehicle}
Yang, C., Li, W., Lin, Z., 2018. Vehicle object detection in remote sensing
  imagery based on multi-perspective convolutional neural network. ISPRS Int.
  J. Geo-Inf 7~(7), 249.

\bibitem[{Zanotta et~al.(2018)Zanotta, Zortea, and Ferreira}]{ZANOTTA2018162}
Zanotta, D.~C., Zortea, M., Ferreira, M.~P., 2018. A supervised approach for
  simultaneous segmentation and classification of remote sensing images. ISPRS
  J. Photogramm. Remote Sens. 142, 162 -- 173.

\end{thebibliography}
\end{document}